\pdfoutput=1
\PassOptionsToPackage{table}{xcolor}
\PassOptionsToPackage{backref=page}{hyperref}
\documentclass[11pt]{article}

\usepackage{EMNLP2022}

\renewcommand*\backref[1]{\ifx#1\relax \else (Cited on p. #1) \fi}

\usepackage{times}
\usepackage{latexsym}

\usepackage[T1]{fontenc}

\usepackage[utf8]{inputenc}

\usepackage{microtype}

\usepackage{inconsolata}

\usepackage{booktabs}
\usepackage{graphicx}
\usepackage{makecell}
\usepackage{amsmath}
\usepackage{amsfonts}
\usepackage{caption}
\usepackage{subcaption}

\usepackage{savesym}
\usepackage{bm}
\usepackage{bbm}
\usepackage{mathtools, nccmath}
\usepackage{amsthm}
\usepackage{amssymb}
\usepackage{multirow}
\usepackage{bbding}
\savesymbol{Square}
\savesymbol{Cross}
\savesymbol{TriangleUp}
\savesymbol{TriangleDown}
\usepackage[geometry]{ifsym}
\usepackage{wasysym}
\usepackage{diagbox}

\usepackage{tikz}
\usepackage{tikz-dependency}
\newcommand*\circled[1]{\tikz[baseline=(char.base)]{
            \node[shape=circle,draw,inner sep=.6pt] (char) {#1};}}
            
\DeclarePairedDelimiterXPP\Expect[2]{\mathbb{E}_{#1}}[]{}{#2}%
\newcommand{\bind}{\bm{\mathbbm{1}}}
\DeclareMathOperator{\bx}{\mathbf{x}}

\newcommand{\rs}[4]{\makecell[cc]{\underset{\scriptstyle \pm #2}{\large #1}\ \big/\underset{\pm \scriptstyle #4}{\large #3}}}
\newcommand{\srs}[2]{\makecell[c]{\underset{\scriptstyle \pm #2\phantom{-}}{{\large #1}\phantom{-}}}}
\newcommand{\wsrs}[2]{\makecell[c]{\underset{\scriptstyle \pm #2}{ \phantom{-}{\large #1}\phantom{-}}}}
\newcommand{\cmbold}[1]{\underline{\mathbf{#1}}}

\usepackage{url}

\usepackage[noabbrev,capitalize,nameinlink]{cleveref}

\usepackage{fontawesome5}

\title{Exploring Predictive Uncertainty and Calibration in NLP:\\ A Study on the Impact of Method \& Data Scarcity}

 \author{Dennis Ulmer\textsuperscript{\faCompass} \hspace{.5em}  Jes Frellsen\textsuperscript{\faRobot} \hspace{.5em}  Christian Hardmeier\textsuperscript{\faCompass} \\
       \textsuperscript{\faCompass}Department of Computer Science, IT University of Copenhagen \\ \textsuperscript{\faRobot}Department of Applied Mathematics \& Computer Science,
Technical University of Denmark \\
        \texttt{dennis.ulmer@mailbox.org}}

\begin{document}
\maketitle

\begin{abstract}
    We investigate the problem of determining the predictive confidence (or, conversely, uncertainty) of a neural classifier through the lens of low-resource languages.
    By training models on sub-sampled datasets in three different languages, we assess the quality of estimates from a wide array of approaches and their dependence on the amount of available data. We find that while approaches based on pre-trained models and ensembles achieve the best results overall, the quality of uncertainty estimates can surprisingly \emph{suffer} with more data. We also perform a qualitative analysis of uncertainties on sequences, discovering that a model's total uncertainty seems to be influenced to a large degree by its data uncertainty, not model uncertainty. All model implementations are open-sourced in a software package.
\end{abstract}

\section{Introduction}



In 1877, Italian astronomer Giovanni Schiaparelli described the existence of ``canals'' on the surface of Mars, a finding that was described by a contemporary as a ``very important and perplexing [problem]'' (\citeauthor{young1895manual}, \citeyear{young1895manual}; p. 355). It later turned out that the structures, originally termed \emph{canali} in Italian, were simply mistranslated, since the word can also refer to (natural) channels of water. By that point however, the possibility of irrigation on the red planet had already sept into popular culture, and is still being referenced to this day.
In the meantime, translation has become a task that is increasingly performed by neural networks, which --- in the face of a word such as \emph{canali} --- might simply fall back on the most likely translation given the training data. And while the error above seems fairly innocuous, there are more safety-critical scenarios in which such ambiguities matter and can potentially have negative real-word consequences.
Besides translation, there also exist other language-based problems in which the uncertainty surrounding a model prediction can convey critical information, such as medical analyses \citep{esteva2019guide}, legal case data \citep{frankenreiter2020computational} or analyzing job applications \citep{zimmermann2016data}. Determining model confidence, or, conversely, uncertainty, consequently is an important mean to instill trust in end users and avert harm \citep{bhatt2021uncertainty, jacovi2021formalizing}.
While there exist many works on images \citep{lakshminarayanan2017simple, snoek2019can} and tabular data \citep{ruhe2019bayesian, ulmer2020trust, malinin2021shifts}, the quality of uncertainty estimates provided by neural networks remains underexplored in Natural Language Processing (NLP). In addition, as model underspecification due to insufficient data presents a risk \citep{d2020underspecification}, the increasing interest in less-researched languages with limited resources raises the question of how reliably uncertain predictions can be identified. This lets us pose the following research questions:
 
\begin{itemize}
    \item[\small \textbf{RQ1}] What are the best approaches in terms of uncertainty quality and calibration?
    \item[\small  \textbf{RQ2}] How are models impacted by the amount of available training data?
    \item[\small  \textbf{RQ3}] What are differences in how the different approaches estimate uncertainty?
\end{itemize}%

\begin{figure}[t]
    \centering 
    \includegraphics[width=0.495\textwidth]{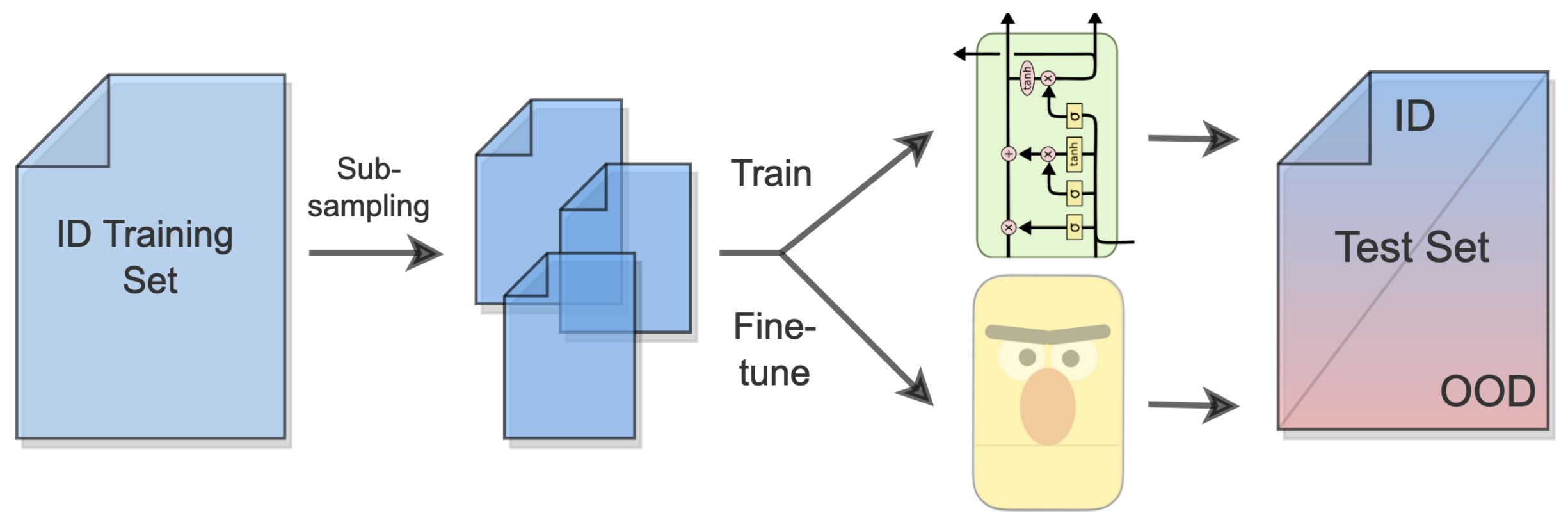}
    \caption{\textbf{Schematic of our experiments.} Training sets are sub-sampled and used to train LSTM-based models and fine-tune transformer-based ones, which are evaluated on in- and out-of-distribution test data.}\label{fig:training}
\end{figure}

 \paragraph{Contributions} \circled{1} We address these questions by conducting a comprehensive empirical study of eight different models for uncertainty estimation for classification and evaluate their effectiveness on three languages spanning distinct NLP tasks, involving sequence labeling and classification. \circled{2} We show that while approaches based on pre-trained models and ensembles achieve the best results overall, the quality of uncertainty estimates on OOD data can become worse using \emph{more} data. \circled{3} In a qualitative analysis, we also discover that a model's total uncertainty seems to mostly consist of its data uncertainty. \circled{4} We make our experimental code and model implementations available open-source in separate repositories, aiding future research in this direction.\footnote{The model zoo is available under \url{https://github.com/Kaleidophon/nlp-uncertainty-zoo}, with the code for the experiments available under \url{https://github.com/Kaleidophon/nlp-low-resource-uncertainty}.}

\section{Related Work}

\paragraph{Notions of Uncertainty} In the absence of additional information, the introductionary example \emph{canali} has two valid translations --- \emph{canals} and \emph{channels}. This is an instance of \emph{data} or \emph{aleatoric} uncertainty, describing the irreducible ambiguity and noise in the data generating process. The other notion is \emph{model} or \emph{epistemic} uncertainty: Fitting parameters, there remains a degree of incertitude about the optimal values due to finite data. We can usually reduce this uncertainty by amassing more data,\footnote{That is, unless the model class we chose is too restrictive.} for instance by supplying a translation system with other meanings of \emph{canali}. These two concepts form the basis for uncertainty estimation in Machine Learning \citep{der2009aleatory, hullermeier2021aleatoric}. 

\paragraph{Uncertainty in NLP} Since uncertainty estimation literature is manifold on image data, we dedicate this part to related works in the realm of Natural Language Processing. There are several examples trying to incorporate uncertainty into models to either increase trustworthiness or performance, for instance in Machine Translation \citep{glushkova2021uncertainty, wei2020uncertainty, xiao2020wat}, Summarization \citep{gidiotis2021uncertainty}, Information Retrieval \citep{penha2021on} and Active Learning \citep{siddhant2018deep}. To obtain uncertainties, \citet{gan2017scalable} use Stochastic-Gradient Langevin Dynamics \citep{welling2011bayesian} to obtain posterior weight samples for a LSTM. \citet{shelmanov2021certain} apply MC Dropout with determinantal point processes to transformers for Natural Language Understanding.
Several authors have also highlighted connections of multi-head attention to Bayesian inference \citep{an2020repulsive, hron2020infinite}. \citet{shen2020modeling} attempt to transfer the idea of prior networks \citep{malinin2018predictive, joo2020being} onto recurrent neural networks. 
Another line of works investigates uncertainty properties themselves; For instance, \citet{chen2022explaining} try to explain uncertainty estimates for BERT and RoBERTa. Another example is given by \citet{xiao2021hallucination}, who use predictive uncertainty to explain hallucination in Language Generation. \citet{xu2020understanding} similarly use uncertainty as a tool to investigate challenges of neural summarization approaches. Lastly, due to the way that uncertainty estimates are evaluated, investigating distributional shift in NLP is also of interest, for instance through the work of \citet{arora2021types}, \citet{kamath2020selective}, who focus on question answering and \citet{tan2019out} for text classification. The most similar work to ours is the text classification uncertainty benchmark by \citet{van2022benchmarking}, however they do not consider the impact of data or language, and test a different selection of models.

\paragraph{Calibration} Calibration denotes the property of a model's output to accurately reflect the true chance of a correct prediction --- i.e. predicting a class with a confidence of $90 \%$ should yield the correct prediction for $90 \%$ of similar inputs, when repeated. There have been several studies testing this property in modern neural networks \citep{guo2017calibration, nixon2019measuring, minderer2021revisiting, wang2021rethinking} and proposing ways to improve it \citep{thulasidasan2019mixup, mukhoti2020calibrating, karandikar2021soft, zhao2021calibrating, tian2021geometric}. In NLP, calibration as been explored for pre-trained models \citep{desai2020calibration}, including on out-of-distribution data \citep{dan2021effects}, for neural machine translation \citep{wang2020on} and for question-answering \citep{jiang2021can}. Likewise, authors have proposed several calibration schemes, for instance by focusing on classes of interest \citep{jagannatha2020calibrating}, generating synthetic examples for regularization \citep{kong2020calibrated}, using richer input representations \citep{zhang2021knowing} and adapting prompts in a zero-shot  setting \citep{zhao2021calibrate}.

\section{Methodology}

\subsection{Models}\label{sec:models}

We choose a variety of models that cover a range of different approaches based on the two most prominently used architectures in NLP: Long-Short Term Memory networks (LSTMs; \citeauthor{hochreiter1997long}, \citeyear{hochreiter1997long}) and transformers \citep{vaswani2017attention}. Inside the first family, we use the Variational LSTM \citep{gal2016theoretically} based on MC Dropout \citep{gal2016dropout}, the Bayesian LSTM \citep{fortunato2017bayesian} implementing Bayes-by-backprop \citep{blundell2015weight} and the ST-$\tau$ LSTM \citep{wang2021uncertainty}, modelling transitions in a finite-state automaton, as well as an ensemble \citep{lakshminarayanan2017simple}. In the second family, we count the Variational Transformer \citep{xiao2020wat}, also using MC Dropout, the SNGP Transformer \citep{liu2022simple}, using a Gaussian Process output layer, and the Deep Deterministic Uncertainty transformer (DDU; \citeauthor{mukhoti2021deterministic}, \citeyear{mukhoti2021deterministic}), fitting a Gaussian mixture model on extracted features. We elaborate on implementation details in \cref{app:implementation-details}.

\subsection{Uncertainty Metrics}\label{sec:metrics}

We employ the following metrics to quantify confidence or uncertainty --- in all cases, lower values indicate lower confidence / certainty and conversely, higher values mean higher confidence / certainty. The following metrics were either chosen due to their frequent use in the literature, or because they are trying to capture uncertainty in a novel way.


\paragraph{Single prediction metrics} We  distinguish between metrics suitable for models using only a single prediction (or using the mean of multiple predictions, e.g. for an ensemble). The most straightforward of them is the maximum softmax probability by \citet{hendrycks2017baseline}. A variant of this is the softmax-gap, measuring the difference between the two largest predicted probabilities \citep{tagasovska2019single}. Another common metric, predictive entropy, involves measuring the Shannon entropy of the output distribution, which is maximized for a uniform prediction:

\begin{equation*}
    -\sum_{k=1}^K p_{\theta}(y=k|\bx) \log p_{\theta}(y=k|\bx)
\end{equation*}

Lastly, we consider the Dempster-Shafer metric \citep{sensoy2019evidential}, defined as $ K / (K + \sum_{k=1}^K \exp(z_k))$, where $z_k$ denotes the logit corresponding to class $k$. It has been shown that probabilities for (ReLU) networks tend to saturate in the limit \citep{hein2019relu, ulmer2021know}, and since this metric considers logits, it might provide more informative estimates on OOD data.

\paragraph{Multiple prediction metrics} For some of the included models, we can express uncertainty as some score based on a number of predicted distributions, e.g. from different ensemble members or forward passes for MC Dropout. Here we use the expectation with respect to the weight posterior to express the aggregation of multiple predictions, which will simply be evaluated using the mean of a number of Monte Carlo samples in practice. A simple uncertainty metric on this basis is the predictive variance between predictions for a class:

\begin{equation*}\begin{aligned}
\resizebox{\hsize}{!}{
    $\displaystyle\frac{1}{K}\sum_{k=1}^K\mathbb{E}_{q(\theta)}\bigg[\bigg(p_{\theta}(y=k|\bx) 
     - \Expect[\Big]{q(\theta)}{p_{\theta}(y=k|\bx)}\bigg)^2\bigg]$
     ,}
\end{aligned}\end{equation*}


where the expectation is evaluated over multiple sets of parameters, e.g.\@ stemming from different dropout masks. Another possibility lies in using the mutual information between the label and model parameters given the data and input sample, which was introduced by \citet{smith2018gal}:

\begin{equation}\label{eq:mutual-information}
    \text{H}\bigg[\Expect[\Big]{q(\theta)}{p_{\theta}(y|\bx)}\bigg] - \Expect[\bigg]{q(\theta)}{\text{H}\Big[p_{\theta}(y|\bx)\Big]}
\end{equation}

where H denotes the Shannon entropy as used for predictive entropy. The two terms of this equation can be identified as the total entropy and the aleatoric uncertainty, respectively. In theory, the remaining epistemic uncertainty of the model --- in form of the the mutual information --- should be particularly high on OOD inputs.

\paragraph{Model-specific metrics} Lastly, DDU by \citet{mukhoti2021deterministic} uses the log-probability of the last layer network activation under a Gaussian Mixture Model fitted on the training set as an additional metric. Since all others models are trained or fine-tuned as classifiers, they are not able to assign log-probabilities to sequences. 

\paragraph{Uncertainty for sequences} Since some tasks require predictions for every time step of a sequence, we determine the uncertainty of a whole sequence in these cases by taking the mean over all step-wise uncertainties.\footnote{We also just considered the \emph{maximum} uncertainty over a sequence, with similar results.} A more principled approach for sequences is for instance provided by \citet{malinin2021uncertainty}, and we leave the extension and exploration of such methods for different uncertainty metrics, models and tasks to future work.

\subsection{Dataset Selection \& Creation}\label{sec:data-creation}

\begin{table*}[ht!]
    \centering 
    \resizebox{0.925\textwidth}{!}{
        \renewcommand{\arraystretch}{2.25}
        \begin{tabular}{@{}llllrr@{}}
            \toprule
            Language & Task & Dataset & OOD Test Set & \makecell[tr]{\# ID / OOD}. & \makecell[tr]{Sub-sampled\\ Training Set Sizes} \\
            \midrule
            EN & \makecell[tl]{Intent\\ Classification} & \makecell[tl]{Clinc Plus  \citep{larson2019evaluation}} & Out-of-scope voice commands & \makecell[tr]{15k / 1k} & \makecell[tr]{15k / 12.5k / 10k}\\
            DA & \makecell[tl]{Named Entity\\ Recognition} & \makecell[tl]{Dan+ News \citep{plank2020dan+}} & Tweets &  \makecell[tr]{4382 / 109} &  \makecell[tr]{4k / 2k / 1k}\\
            FI & PoS Tagging & \makecell[tl]{Finnish UD Treebank  (\citeauthor{haverinen2013tdt}, \citeyear{haverinen2013tdt};\\  \citeauthor{pyysalo2015udfinnish}, \citeyear{pyysalo2015udfinnish};
            \citeauthor{kanerva-2022-ood}, \citeyear{kanerva-2022-ood})} & \makecell[tl]{Hospital records, online forums,\\ tweets, poetry} & \makecell[tr]{12217 / 2122} &  \makecell[tr]{10k / 7.5k / 5k}\\
            \bottomrule
        \end{tabular}%
    }
    \caption{\textbf{Datasets}. The original and sub-sampled number of sequences for experiments are given on the right.}\label{table:datasets}
\end{table*}


\paragraph{In-distribution training sets} We choose three different languages, namely English (Clinc Plus; \citeauthor{larson2019evaluation}, \citeyear{larson2019evaluation}), Danish in the form of the Dan+ dataset \citep{plank2020dan+} based on News texts from PAROLE-DK
\citep{bilgram1998construction}, Finnish (UD Treebank; \citeauthor{haverinen2013tdt}, \citeyear{haverinen2013tdt}; \citeauthor{pyysalo2015udfinnish}, \citeyear{pyysalo2015udfinnish}; \citeauthor{kanerva-2022-ood}, \citeyear{kanerva-2022-ood}), corresponding to NLP tasks such as sequence classification, named entity recognition and part-of-speech tagging. An overview over the used the data is given in \cref{table:datasets}. We do use standardized low-resource languages in the case of Finnish and Danish, and simulate a low-resource setting using English data.\footnote{The definition of low-resource actually differs greatly between works. One definition by \citet{bird2022local} advocates the usage for (would-be) standardized languages with a large amount of speakers and a written tradition, but a lack of resources for language technologies. Another way is a task-dependent definition: For dependency parsing, \citet{eberstein2021genre} define low-resource as providing less than $5000$ annotated sentences in the Universal Dependencies Treebank. \citet{hedderich2021survey,lignos2022toward} lay out a task-dependent spectrum, from a several hundred to thousands of instances.} Starting with a sufficiently-sized training set and then sub-sampling allows us to create training sets of arbitrary sizes. By using languages from different families, we hope to be able draw conclusions that generalize across a single language. We employ a specific sampling scheme that tries to maintain the sequence length and class distribution of the original corpus, which we explain and verify in \cref{app:training-set}.\\

\paragraph{Out-of-distribution Test Sets} While it is possible to create OOD text by for instance withholding classes from the training set or appending text from a different source \citep{arora2021types}, we choose to pick entirely new OOD test sets that are qualitatively different: Out-of-scope voice commands by users in \citet{larson2019evaluation},\footnote{Since all instances in this test set correspond to out-of-scope inputs and not to classes the model was trained on, we cannot evaluate certain metrics in \cref{table:results}.} the Twitter split of the Dan+ dataset \citep{plank2020dan+}, and the Finnish OOD treebank \citep{kanerva-2022-ood}. In similar works for the image domain, OOD test sets are often chosen to be convincingly different from the training distribution, for instance MNIST versus Fashion-MNIST \citep{nalisnick2019do,van2021feature}. While there exist a variety of formalizations of types of distributional shift \citep{moreno2012unifying,wald2021calibration,arora2021types,federici2021information}, it is often hard to determine if and what kind of shift is taking place.
\citet{winkens2020contrastive} define \emph{near OOD} as a scenario in which the inlier and outlier distribution are meaningfully related, and \emph{far OOD} as a case in which they are unrelated. Unfortunately, this distinction is somewhat arbitrary and hard to apply in a language context, where OOD \emph{could} be defined as anything ranging from a different language or dialect to a different demographic on an author or speaker or a new genre.
Therefore, we use a similar methodology to the validation of the sub-sampled training sets to make an argument that the selected OOD splits are sufficiently different in nature from the training splits. The exact procedure along some more detailed results is described in \cref{app:ood-test-set}.

\subsection{Model Training}\label{sec:model-training}

Unfortunately, our datasets do not contain enough data to train transformer-based models from scratch. Therefore, we only fully train LSTM-based models, while using pre-trained transformers, namely BERT (English; \citeauthor{devlin2019bert}, \citeyear{devlin2019bert}), Danish BERT \citep{hvingelby2020dane}, and FinBERT (Finnish; \citeauthor{virtanen2019multilingual}, \citeyear{virtanen2019multilingual}), for the other approaches. The whole procedure is depicted in \cref{fig:training}. The way we optimize models is provided in \cref{app:transformer-fine-tuning}. We list training hardware, hyperparameter information in \cref{app:training-details}, with the environmental impact described in \cref{sec:environmental-impact}.

\subsection{Evaluation}\label{sec:evaluation}

Apart from evaluating models on the task performance, we also evaluate the following calibration and uncertainty, painting a multi-faceted picture of the reliability of models. In all cases, we use the Almost Stochastic Order test (ASO; \citeauthor{del2018optimal}, \citeyear{del2018optimal}; \citeauthor{dror2019deep}, \citeyear{dror2019deep}) for significance testing, which is elaborated on in \cref{app:implementation-details}.

\paragraph{Evaluation of Calibration} First, we measure the calibration of models using the adaptive calibration error (ACE; \citeauthor{nixon2019measuring}, \citeyear{nixon2019measuring}), which is an extension of the expected calibration error (ECE; \citeauthor{naeini2015obtaining}, \citeyear{naeini2015obtaining}; \citeauthor{guo2017calibration}, \citeyear{guo2017calibration}).\footnote{See \cref{app:calibration-metrics} for a short overview over the differences.} Furthermore, we use the frequentist measure of coverage \citep{larry2004all, kompa2021empirical}. Coverage is based on the prediction set $\hat{\mathbb{P}}(\mathbf{x})$ of a classifier given an input, which includes the most likely classes adding up to or surpassing $1 - \alpha$ probability mass. A well-tuned classifier should contain the correct class in this very set, and minimize its width. The extent to which this property holds can be determined by the \emph{coverage percentage}, i.e.\@ the number of times the correct class in indeed contain within the prediction set, and its cardinality, denoted simply as \emph{width}. 

\paragraph{Evaluation of Uncertainty} We compare uncertainty scores on the ID and OOD test set and measure the area under the receiver-operator curve (AUROC) and under the precision-recall curve (AUPR), assuming that uncertainty will generally be higher on samples from the OOD test set.\footnote{We thus formulate a pseudo-binary classification task as common in the literature, using the model's uncertainty score to try to distinguish the two test sets. Note that we do not advocate for actually using uncertainty for OOD detection, but only use it for evaluation purposes, since uncertainty on OOD examples should be high due to model uncertainty.} An ideal model should create very different distributions of confidence scores on ID and OOD data, thus maximizing AUROC and AUPR. However, we also want to find out to what extend uncertainty can give an indication of the correctness of the model, which is why we propose a new way to evaluate the \emph{discrimination} property posed by \citet{alaa2020discriminative} based on \citet{leonard1992neural}: A good model should be less certain for inputs that incur a higher loss. To measure this both on a token and sequence level, we utilize Kendall's $\tau$ \citep{kendall1938new}, which, given two lists of measurements, determines the degree to which they are \emph{concordant} --- that is, to what extent the rankings of elements according to their measured values agree. This is expressed by a value between $-1$ and $1$, with the latter expressing complete concordance. In our case, these measurements correspond to the uncertainty estimate and the actual model loss, either for tokens (Token $\tau$) or sequences (Sequence $\tau$).

\section{Experiments}\label{sec:experiments}


\subsection{RQ1: Uncertainty \& Calibration}\label{sec:uncertainty-calibration}

\begin{table*}[ht!]
    \centering 
    \resizebox{0.9975\textwidth}{!}{
        \renewcommand{\arraystretch}{2}
        \rowcolors{2}{gray!15}{white}
        \begin{tabular}{@{}ll||cc|cccc|cccc@{}}
            \toprule
             & & \multicolumn{2}{c|}{Task $\big( \text{ID} \big/ \text{OOD}\big)$} &  \multicolumn{4}{c|}{Calibration $\big( \text{ID} \big/ \text{OOD}\big)$} & \multicolumn{4}{c}{Uncertainty $\big( \text{ID} \big/ \text{OOD}\big)$} \\
           & Model & Acc.$\uparrow$ & $F_1\uparrow$ & ECE$\downarrow$ & ACE$\downarrow$ & \%\@Cov.$\uparrow$ &  $\varnothing$Width$\downarrow$ & AUROC$\uparrow$ & AUPR$\uparrow$ & Token $\tau\uparrow$ & Seq. $\tau\uparrow$ \\
            \midrule
             & LSTM & $\wsrs{.79}{.00}$ & $\wsrs{.62}{.01}$ & $\wsrs{77.95}{.00}$  & $\wsrs{.49}{.01}$ & $\wsrs{\cmbold{1.00}}{.00}$ & $\wsrs{144.00}{.00}$ & $\srs{.88^{\text{\tiny \Plus}}}{.01}$ & $\srs{.60^{\text{\tiny \Plus}}}{.01}$ & \backslashbox[15mm]{}{} & $\wsrs{.75^{\bigcirc}}{.01}$  \\
             & Bayesian LSTM & $\wsrs{.59}{.06}$ & $\wsrs{.46}{.05}$ & $\wsrs{77.66}{.05}$ &  $\wsrs{.22}{.01}$ & $\wsrs{.88}{.00}$ & $\srs{41.99}{1.94}$ & $\srs{.86^{\bigtriangleup}}{.01}$ & $\srs{.59^{\text{\tiny \XSolidBold}}}{.01}$ & \backslashbox[15mm]{}{} & $\wsrs{.66^{\bigcirc}}{.02}$  \\
             & LSTM Ensemble & $\wsrs{\cmbold{.81}}{.00}$ & $\wsrs{\cmbold{.64}}{.01}$ & $\wsrs{77.11}{.00}$ &  $\wsrs{.09}{.00}$ & $\wsrs{.87}{.00}$ & $\wsrs{4.27}{.05}$ & $\srs{\cmbold{.92^{\text{\tiny \Plus}}}}{.00}$ & $\srs{\cmbold{.71^{\text{\tiny \Plus}}}}{.01}$ & \backslashbox[15mm]{}{} & $\wsrs{.73^{\Box}}{.01}$  \\
             & Variational BERT & $\wsrs{.45}{.16}$ & $\wsrs{.34}{.13}$ & $\wsrs{77.92}{.02}$ &  $\wsrs{.22}{.06}$ & $\wsrs{1.00}{.00}$ & $\wsrs{115.11}{11.38}$ & $\srs{.80^{\text{\tiny \XSolidBold}}}{.01}$ & $\srs{.53^{\text{\tiny \XSolidBold}}}{.01}$ & \backslashbox[15mm]{}{} & $\wsrs{.57^{\bigcirc}}{.09}$ \\ 
            \multirow{-5}{*}{\rotatebox{90}{\textbf{English}}}  & DDU BERT & $\wsrs{.79}{.00}$ & $\wsrs{.64}{.01}$ & $\wsrs{\cmbold{77.02}}{.00}$ &  $\wsrs{\cmbold{.00}}{.00}$ & $\wsrs{.82}{.00}$ & $\wsrs{\cmbold{1.46}}{.04}$ & $\srs{.88^{\bigcirc}}{.00}$ & $\srs{.62^{\bigcirc}}{.01}$ & \backslashbox[15mm]{}{} & $\wsrs{\cmbold{.87^{\bigcirc}}}{.00}$  \\[0.1cm]
            \midrule
             & LSTM & $\rs{.93}{.00}{.92}{.00}$ & $\rs{.26}{.01}{.19}{.01}$ & $\rs{17.18}{.00}{17.17}{.00}$ &  $\rs{.16}{.01}{.10}{.01}$ & $\rs{\cmbold{1.00}}{.00}{\cmbold{1.00}}{.00}$ & $\rs{19.00}{.00}{19.00}{.00}$ & $\srs{.50^{\bigcirc}}{.02}$ & $\srs{.14^{\bigcirc}}{.01}$ & $\rs{.50^{\bigcirc}}{.01}{.47^{\bigcirc}}{.00}$ & $\rs{-.26^{\text{\tiny \Plus}}}{.02}{-.28^{\bigcirc}}{.05}$  \\
             & Variational LSTM & $\rs{.90}{.02}{.90}{.02}$ & $\rs{.08}{.02}{.09}{.02}$ & $\rs{16.74}{.03}{16.72}{.03}$ & $\rs{.26}{.02}{.17}{.01}$ & $\rs{.99}{.01}{.98}{.01}$ & $\rs{6.62}{.37}{6.68}{.33}$ & $\srs{.60^{\text{\tiny \Plus}}}{.04}$ & $\srs{.21^{\text{\tiny \Plus}}}{.02}$ & $\rs{.23^{\bigcirc}}{.06}{.23^{\bigcirc}}{.05}$ & $\rs{-.04^{\text{\tiny \XSolidBold}}}{.02}{-.02^{\Box}}{.05}$  \\
             & ST-$\tau$ LSTM & $\rs{.92}{.00}{.92}{.00}$ & $\rs{.12}{.00}{.09}{.00}$ & $\rs{16.67}{.00}{16.63}{.01}$ &  $\rs{.24}{.01}{.15}{.01}$ & $\rs{1.00}{.00}{.99}{.00}$ & $\rs{7.10}{.07}{7.03}{.08}$ & $\srs{.54^{\text{\tiny \Plus}}}{.01}$ & $\srs{.15^{\text{\tiny \Plus}}}{.01}$ & $\rs{.50^{\bigcirc}}{.00}{.48^{\bigcirc}}{.00}$ & $\rs{-.05^{\Box}}{.03}{-.01^{\Box}}{.05}$  \\
             & Bayesian LSTM & $\rs{.93}{.00}{.93}{.00}$ & $\rs{.07}{.00}{.07}{.00}$ & $\rs{16.81}{.00}{16.79}{.00}$ &  $\rs{.25}{.01}{.18}{.01}$ & $\rs{1.00}{.00}{1.00}{.00}$ & $\rs{1.68}{.04}{1.70}{.05}$ & $\srs{.65^{\pentagon}}{.17}$ & $\srs{.31^{\pentagon}}{.30}$ & $\rs{.53^{\bigcirc}}{.01}{\cmbold{.55^{\bigcirc}}}{.01}$ & $\rs{-.01^{\Box}}{.07}{-.02^{\text{\tiny \Plus}}}{.04}$  \\
             & LSTM Ensemble & $\rs{\cmbold{.95}}{.00}{\cmbold{.94}}{.00}$ & $\rs{\cmbold{.33}}{.01}{\cmbold{.25}}{.01}$ & $\rs{16.37}{.00}{\cmbold{16.35}}{.00}$ & $\rs{.18}{.01}{.13}{.01}$ & $\rs{.98}{.00}{.97}{.00}$ & $\rs{\cmbold{1.62}}{.00}{\cmbold{1.58}}{.01}$ & $\srs{.60^{\Box}}{.02}$ & $\srs{.18^{\Box}}{.01}$ & $\rs{.44^{\Box}}{.00}{.45^{\Box}}{.00}$ & $\rs{-.19^{\text{\tiny \Plus}}}{.01}{-.28^{\Box}}{.01}$  \\
             & SNGP BERT & $\rs{.22}{.35}{.19}{.34}$ & $\rs{.03}{.03}{.02}{.02}$ & $\rs{17.19}{.01}{17.18}{.01}$ &  $\rs{\cmbold{.08}}{.01}{\cmbold{.06}}{.01}$ & $\rs{1.00}{.00}{1.00}{.00}$ & $\rs{18.84}{.32}{18.83}{.34}$ & $\srs{.86^{\bigtriangleup}}{.06}$ & $\srs{.49^{\bigtriangleup}}{.12}$ & $\rs{.17^{\Box}}{.09}{.26^{\Box}}{.14}$ & $\rs{\cmbold{.29^{\text{\tiny \XSolidBold}}}}{.03}{\cmbold{.44^{\Box}}}{.11}$  \\
             & Variational BERT & $\rs{.94}{.00}{.89}{.00}$ & $\rs{.29}{.01}{.17}{.00}$  & $\rs{\cmbold{16.36}}{.00}{16.43}{.00}$ &  $\rs{.20}{.00}{.22}{.00}$ & $\rs{.99}{.00}{.98}{.00}$ & $\rs{2.25}{.01}{3.86}{.08}$ & $\srs{.86^{\text{\tiny \Plus}}}{.01}$ & $\srs{.46^{\text{\tiny \Plus}}}{.02}$ & $\rs{.42^{\bigcirc}}{.00}{.17^{\pentagon}}{.00}$ & $\rs{-.35^{\Box}}{.01}{-.41^{\Box}}{.01}$  \\
            \multirow{-8}{*}{\rotatebox{90}{\textbf{Danish}}}  & DDU BERT & $\rs{.92}{.00}{.89}{.00}$ & $\rs{.25}{.00}{.17}{.00}$ & $\rs{16.41}{.00}{16.44}{.00}$ &  $\rs{.19}{.01}{.21}{.01}$ & $\rs{.99}{.00}{.99}{.00}$ & $\rs{3.48}{.01}{4.04}{.03}$ & $\srs{.86^{\bigcirc}}{.01}$ & $\srs{.39^{\bigcirc}}{.02}$ & $\rs{\cmbold{.56^{\bigcirc}}}{.00}{.25^{\bigcirc}}{.01}$ & $\rs{-.24^{\bigcirc}}{.01}{-.38^{\bigcirc}}{.03}$  \\[0.1cm]
            \midrule
             & LSTM & $\rs{.75}{.00}{.69}{.00}$ & $\rs{.57}{.00}{.53}{.00}$ & $\rs{6.78}{.00}{6.80}{.00}$ & $\rs{.40}{.01}{.38}{.01}$ & $\rs{1.00}{.00}{1.00}{.00}$ & $\rs{16.00}{.00}{16.00}{.00}$ & $\srs{.63^{\bigtriangleup}}{.01}$ & $\srs{.69^{\text{\tiny \Plus}}}{.01}$ & $\rs{.29^{\bigcirc}}{.00}{.19^{\bigcirc}}{.01}$ & $\rs{-.28^{\text{\tiny \Plus}}}{.02}{-.27^{\text{\tiny \Plus}}}{.02}$  \\
             & Variational LSTM & $\rs{.27}{.00}{.26}{.00}$ & $\rs{.03}{.00}{.03}{.00}$ & $\rs{6.65}{.01}{6.66}{.01}$ &  $\rs{.27}{.01}{.28}{.01}$ & $\rs{.97}{.00}{.96}{.00}$ & $\rs{1.35}{.23}{1.37}{.21}$ & $\srs{.51^{\text{\tiny \Plus}}}{.01}$ & $\srs{.59^{\text{\tiny \Plus}}}{.01}$ & $\rs{.00^{\bigtriangleup}}{.01}{.00^{\pentagon}}{.00}$ & $\rs{.01^{\bigtriangleup}}{.03}{.01^{\Box}}{.01}$  \\
             & ST-$\tau$ LSTM & $\rs{.76}{.00}{.71}{.00}$ & $\rs{.58}{.00}{.55}{.00}$ & $\rs{6.18}{.00}{6.21}{.00}$ &  $\rs{.20}{.01}{.22}{.01}$ & $\rs{.97}{.00}{.96}{.00}$ & $\rs{3.32}{.01}{3.57}{.01}$ & $\srs{.62^{\bigtriangleup}}{.01}$ & $\srs{.69^{\text{\tiny \Plus}}}{.01}$ & $\rs{.31^{\bigcirc}}{.00}{.21^{\bigcirc}}{.01}$ & $\rs{-.14^{\text{\tiny \Plus}}}{.02}{-.12^{\Box}}{.04}$  \\
             & Bayesian LSTM & $\rs{.27}{.00}{.26}{.00}$ & $\rs{.03}{.00}{.03}{.00}$ & $\rs{6.84}{.00}{6.85}{.00}$ &  $\rs{\cmbold{.11}}{.00}{\cmbold{.12}}{.00}$ & $\rs{1.00}{.00}{1.00}{.00}$ & $\rs{16.00}{.00}{16.00}{.00}$ & $\srs{.51^{\pentagon}}{.01}$ & $\srs{.60^{\text{\tiny \XSolidBold}}}{.00}$ & $\rs{.00^{\pentagon}}{.00}{.00^{\pentagon}}{.00}$ & $\rs{.01^{\bigcirc}}{.01}{.04^{\text{\tiny \Plus}}}{.00}$  \\
             & LSTM Ensemble & $\rs{.81}{.00}{.75}{.00}$ & $\rs{.62}{.00}{.57}{.00}$ & $\rs{6.18}{.00}{6.22}{.00}$ &  $\rs{.17}{.01}{.21}{.00}$ & $\rs{.99}{.00}{.98}{.00}$ & $\rs{3.46}{.01}{3.80}{.01}$ & $\srs{\cmbold{.67^{\text{\tiny \Plus}}}}{.01}$ & $\srs{\cmbold{.74^{\text{\tiny \Plus}}}}{.01}$ & $\rs{.29^{\bigcirc}}{.00}{.19^{\bigcirc}}{.01}$ & $\rs{-.28^{\text{\tiny \Plus}}}{.01}{-.31^{\text{\tiny \Plus}}}{.01}$  \\
             & Variational BERT & $\rs{.87}{.00}{.81}{.00}$ & $\rs{.74}{.00}{.70}{.00}$ & $\rs{6.11}{.00}{6.15}{.00}$ &  $\rs{.14}{.00}{.18}{.01}$ & $\rs{.99}{.00}{.99}{.00}$ & $\rs{4.68}{.03}{5.19}{.02}$ & $\srs{.64^{\bigtriangleup}}{.01}$ & $\srs{.70^{\bigcirc}}{.01}$ & $\rs{.14^{\bigcirc}}{.00}{.08^{\text{\tiny \Plus}}}{.00}$ & $\rs{-.19^{\text{\tiny \XSolidBold}}}{.00}{-.16^{\text{\tiny \XSolidBold}}}{.01}$  \\
             & SNGP BERT & $\rs{.18}{.10}{.17}{.10}$ & $\rs{.07}{.02}{.08}{.02}$ & $\rs{6.82}{.00}{6.83}{.00}$ &  $\rs{.16}{.02}{.15}{.01}$ & $\rs{1.00}{.00}{.99}{.01}$ & $\rs{15.00}{.00}{15.00}{.00}$ & $\srs{.54^{\bigtriangleup}}{.05}$ & $\srs{.63^{\bigtriangleup}}{.04}$ & $\rs{.15^{\Box}}{.04}{.15^{\Box}}{.03}$ & $\rs{\cmbold{.12^{\Box}}}{.05}{\cmbold{.14^{\Box}}}{.02}$ \\
            \multirow{-8}{*}{\rotatebox{90}{\textbf{Finnish}}}  & DDU BERT & $\rs{.87}{.00}{.81}{.00}$ & $\rs{.72}{.03}{.68}{.03}$ & $\rs{\cmbold{6.01}}{.00}{\cmbold{6.03}}{.00}$ &  $\rs{.33}{.02}{.38}{.02}$ & $\rs{.94}{.00}{.91}{.00}$ & $\rs{\cmbold{2.16}}{.06}{\cmbold{2.31}}{.06}$ & $\srs{.61^{\bigcirc}}{.02}$ & $\srs{.69^{\bigcirc}}{.02}$ & $\rs{\cmbold{.39^{\bigcirc}}}{.04}{\cmbold{.26^{\bigcirc}}}{.03}$ & $\rs{-.07^{\bigcirc}}{.05}{-.16^{\bigcirc}}{.04}$  \\
            \bottomrule
        \end{tabular}%
        }
     \caption{\footnotesize \textbf{\footnotesize Results on the tested datasets.} Task performance is measured by macro $F_1$ and accuracy, calibration by different calibration errors, the coverage percentage the average prediction set width. For every result, and value on the ID and OOD test set is shown. For English, OOD scores are not available since the OOD set does not contain gold labels, and Token $\tau$ is missing due to CLINC being a sequence prediction task. Uncertainty quality is evaluated using its ability to discriminate between ID and OOD data, quantified by AUROC and AUPR. Furthermore, Kendall's $\tau$ is measured between the uncertainty and losses on a sequence- and token-level. Displayed are mean and standard deviation over five random seeds, with bolding and underlining indicating almost stochastic dominance with $\varepsilon_\text{min} \le 0.3$ over all other models. For last section, the best value over uncertainty metrics is given, with symbols indicating the type of metric achieving it: ${\bigcirc}$ Max. probability, ${\bigtriangleup}$ Predictive entropy. ${\Box}$ Class variance. ${\pentagon}$ Softmax gap. ${\text{\tiny \Plus}}$ Dempster-Shafer. ${\text{\tiny \XSolidBold}}$ Mutual information.}\label{table:results}
\end{table*}

We present the results from our experiments using the largest training set sizes per dataset in \cref{table:results}.\footnote{For English, some models were omitted due to convergence issues, which are discussed in \cref{app:convergence-clinc-plus}.} 

\paragraph{Task Performance} Across datasets and models, we can identify several trends: some of the BERT-based models unsurprisingly perform better than LSTM based models, which can be explained with their pre-training procedure. We observe worse performance for some LSTM and BERT-variants, in particular the Variational, Bayesian and ST-$\tau$ LSTM, as well the SNGP BERT. In accordance with the ML literature (see e.g.\@ \citet{lakshminarayanan2017simple, ovadia2019can}, LSTM ensembles actually perform very strongly and on par or sometimes better than fine-tuned BERTs.

\paragraph{Calibration} We also see BERT models to generally achieve lower calibration errors across all metrics measured, which is in line with previous works \citep{desai2020calibration,dan2021effects}. It is interesting to see that the correct prediction is almost always contained in the $0.95$ confidence set across all models, however these number have to be interpreted in the context of the set's width: It becomes apparent that for instance LSTMs achieve this coverage by spreading probability mass over many classes, while only BERT-based models, LSTM ensembles as well as the Bayesian LSTM (on Danish) and the Variational LSTM (on Finnish) are \emph{confidently} correct. 

\paragraph{Uncertainty Quality} LSTM-based model seem to struggle to distinguish in- from out-of-distribution data based on predictive uncertainty. For Danish, only BERTs perform visibly above chance-level. For Finnish, the AUPR results suggest that although some OOD instances are quickly identified as uncertain, many other OOD remain undetected among in-distribution samples. For English, OOD samples are detected more effectively, which can be explained by them consisting of unknown voice commands, representating a potential instance of \emph{semantic} shift, which has been shown to be easier to detect by classifiers \citep{arora2021types}. Furthermore, it is striking that uncertainty and loss on a token-level (Token $\tau$) is only positive correlated for some models, using metrics such as the maximum probability score, softmax gap or the Dempster-Shafer metric, which are all entirely based on the categorical output distributions. On a sequence-level (Sequence $\tau$), the correlation is often \emph{negative}, meaning that higher uncertainty goes hand in hand with a \emph{higher} loss. Lastly, it should be noted that different uncertainty metrics yield diverse outcomes: There does not seem to be one superior metric across all experimental settings, as seen by the variety of markers shown in \cref{table:results}.

\subsection{RQ2: Dependence on training data}\label{sec:dependence-training-data}

\begin{figure*}[ht!]
    \centering
    \includegraphics[width=2\columnwidth]{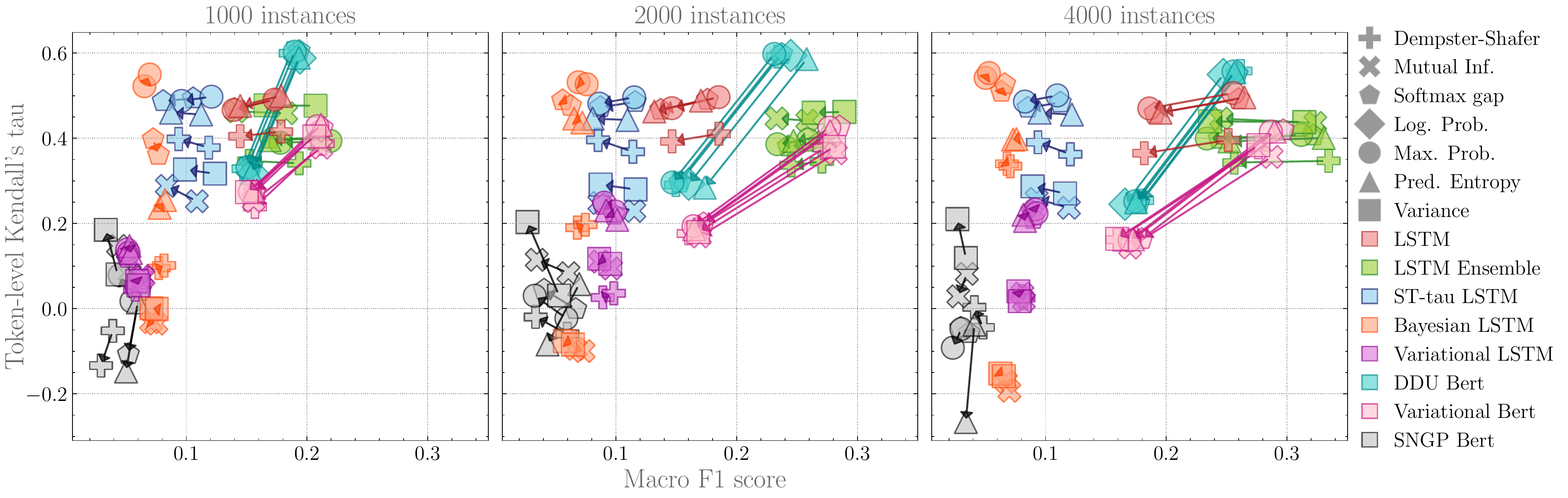}
    \caption{\textbf{Scatter plot showing the difference between model performance (measured by macro $F_1$  and the quality of uncertainty estimates on a token-level (measured by Kendall's $\tau$)}. Shown are different models and uncertainty metrics and several training set sizes of the Dan+ dataset. Arrows indicate changes between the in-distribution and out-of-distribution test set. Best viewed electronically and in color.}\label{fig:scatter-plot-danplus-kendalls-tau-token}
\end{figure*}

After presenting the best results for the biggest training set sizes in \cref{table:results}, we now continue to analyze the difference between models and metrics in a more fine-grained way. 
In \cref{fig:scatter-plot-danplus-kendalls-tau-token}, we show differences for the token-level correlation between a model's loss and its uncertainty measured by Kendall's $\tau$, with arrows indicating the shift from measurements on the in- to the out-of-distribution test set. Here, we see the same trend of more training data having a larger influence on BERT models. Peculiarly, we also observe pre-trained models' uncertainty to correlate less with their losses on the OOD data, while this property stays relative constant for LSTMs. We can recognize this trend also for the other datasets in \cref{fig:scatter-plot-danplus-kendalls-tau-token} and to a lesser degree on a sequence level \cref{subfig:clinc-plus-scatter-kendalls-tau-seq} in \cref{app:additional-scatters}, albeit with a \emph{negative} correlation in general in the latter case.
In \cref{fig:scatter-plot-auroc,fig:scatter-plot-aupr} in \cref{app:additional-scatters}, we show the AUROC and AUPR of different model-uncertainty metric combinations for all datasets and training set sizes. In both cases, we can notice that pre-trained models profit more from an increase in available training data than LSTM-based models that are trained from scratch. This improvement is observed both in task performance, as well as in the model's ability to discern ID from OOD data using its uncertainty, but more so for the Danish than English or Finnish. Like in the previous section, we often see that uncertainty metrics of the same model perform quite similarly. 
These results outline a seeming paradox: Pre-trained and then fine-tuned models (often) perform better on the task at hand, and provide better uncertainty estimates, but only on in-distribution data. Models trained from scratch that have seen less data overall, however provide more reliable uncertainty estimates on OOD data, but are also worse calibrated (\cref{sec:uncertainty-calibration}), with the exception of ensembles. This effect appears to largest on Danish, containing the least data. 

\begin{figure*}[ht]
    \centering
    \begin{subfigure}[t]{0.99\columnwidth}
        \centering
        \includegraphics[width=\columnwidth]{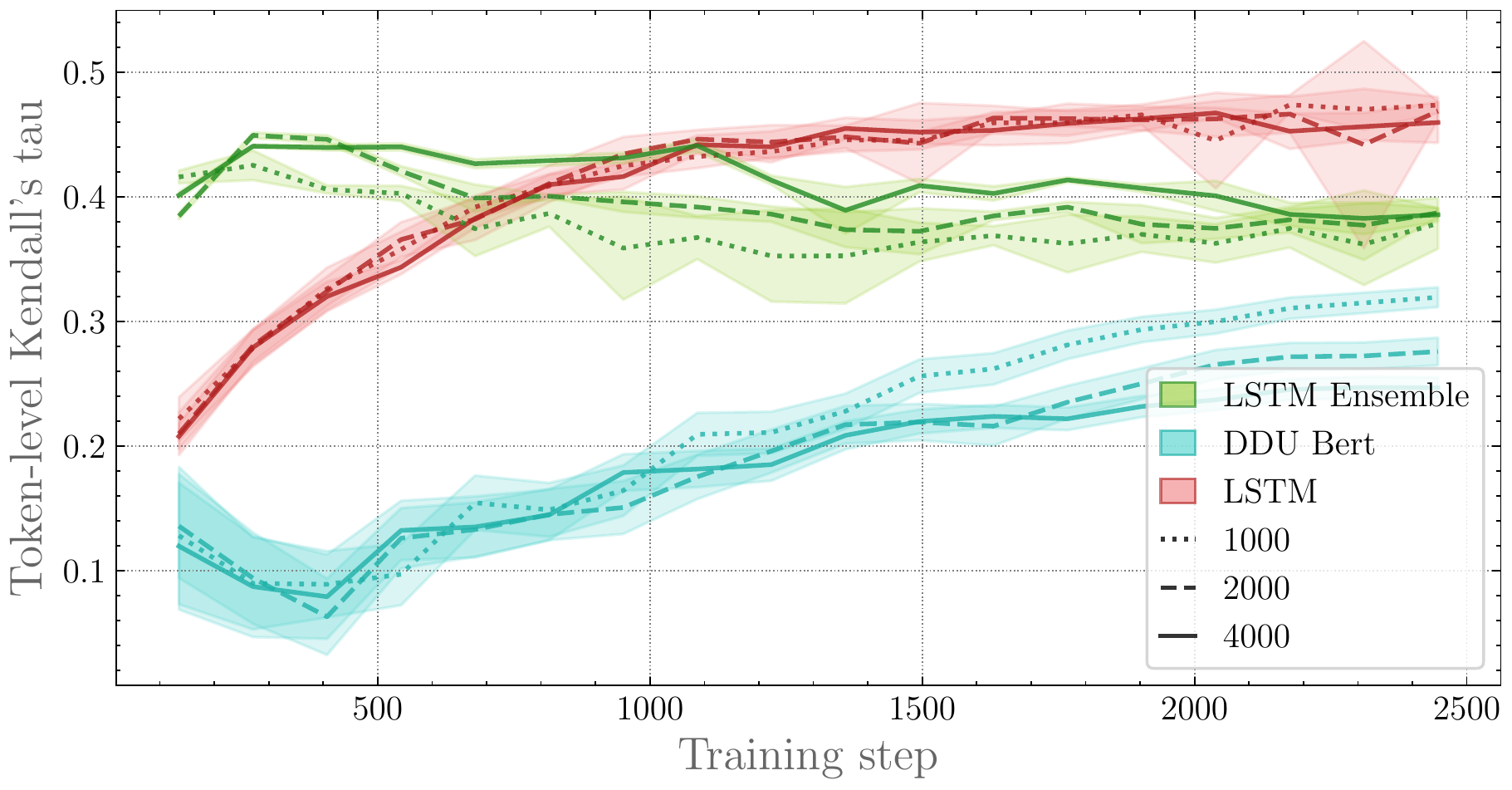}
        \subcaption{Development of token-level Kendall's $\tau$.}
    \end{subfigure}\label{subfig:development-lstm-ddu-predictive-entropy-token}
    \hfill
    \begin{subfigure}[t]{0.99\columnwidth}
        \centering
        \includegraphics[width=\columnwidth]{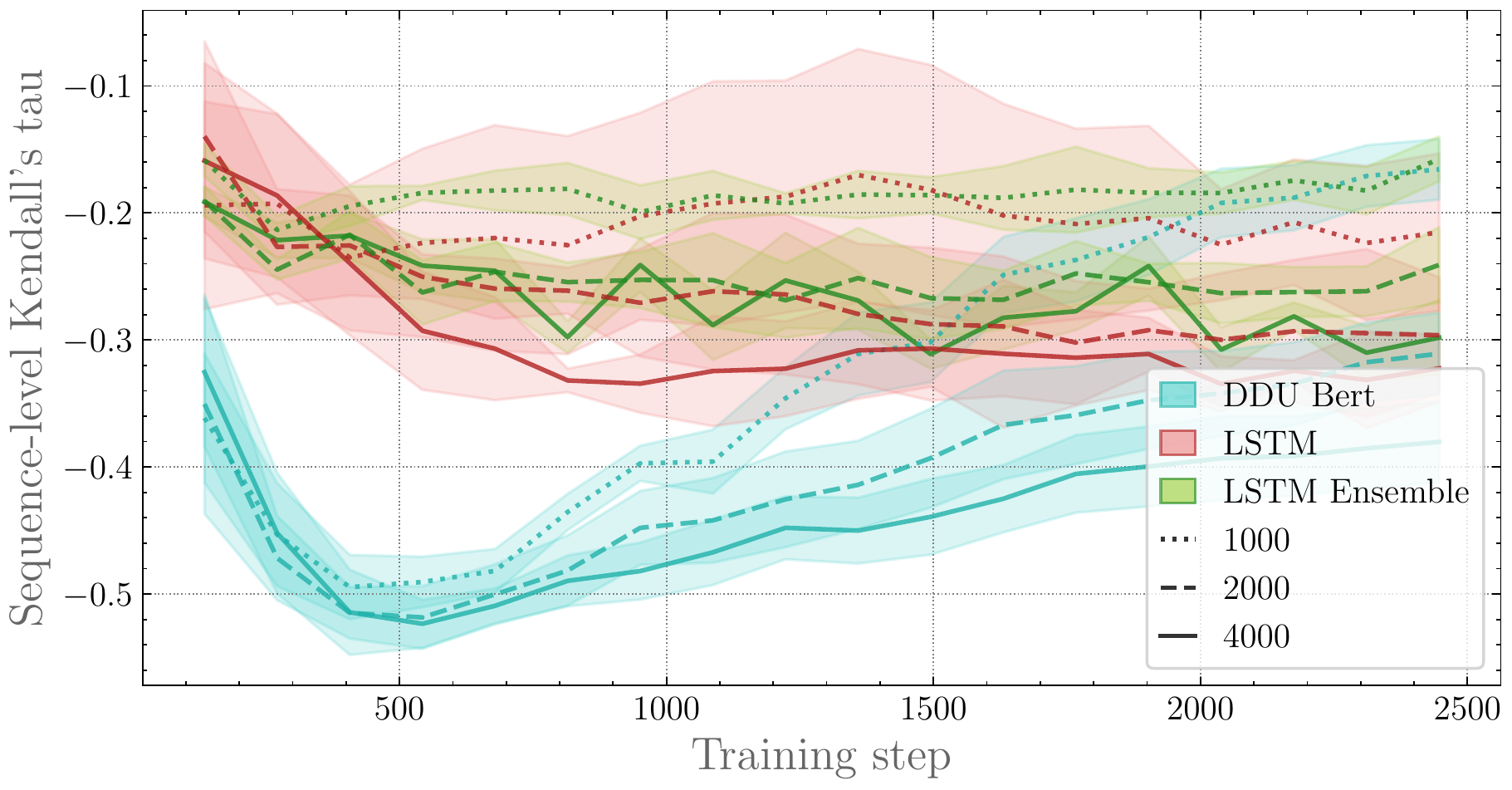}
        \subcaption{Development of sequence-level Kendall's $\tau$.}\label{subfig:development-lstm-ddu-predictive-entropy-seq}
    \end{subfigure}
    \caption{\textbf{Development of correlation between uncertainty and loss}, shown on the Dan+ OOD test set over the training time using differently-sized training sets. Colored areas indicate the standard deviation over five runs.}\label{fig:development-lstm-ddu-predictive-entropy}
\end{figure*}

\paragraph{Uncertainty quality over training} Adding another facet to this issue, we plot the development of uncertainty estimate quality over the training for different models in \cref{fig:development-lstm-ddu-predictive-entropy}. We use LSTMs and DDU BERTs on Dan+ as representative for the observed differences between pre-trained transformers and models trained from scratch, with more examples given in \cref{app:more-uncertainty-over-training}. On both a token and sequence-level, we can see that the correlation between uncertainty and loss dips for DDU BERT, before increasing again over the course the of the training.\footnote{Note that the OOD data used to create these results were not used for training.} Most curiously, the highest correlations are achieved with the models using the \emph{least} training data. Such behavior is also present for LSTMs on a sequence level. We can also see that while the correlation is higher for DDU BERT on in-distribution data (see again \cref{table:results}), on OOD data, LSTMs actually more accurately reflect their knowledge using uncertainty. This again corroborates earlier insights from \cref{sec:uncertainty-calibration}: Pre-trained models seem to provide better uncertainty estimates on in-distribution data, but yield worse results on OOD than LSTMs trained from scratch. Furthermore, the less training data is available, the more indicative predictive uncertainty seems to be of the correctness of a model. We see such behavior also to a lesser extent in the other, datasets (see \cref{app:more-uncertainty-over-training}). Before we offer some potential explanations of this behavior, we try to gain an even more fine-grained understanding by analysing the differences in metrics and models on a token-level.

\subsection{RQ3: Qualitative Analysis}

\begin{figure*}[ht]
    \centering
    \begin{subfigure}[t]{0.99\columnwidth}
        \centering
        \includegraphics[width=\columnwidth]{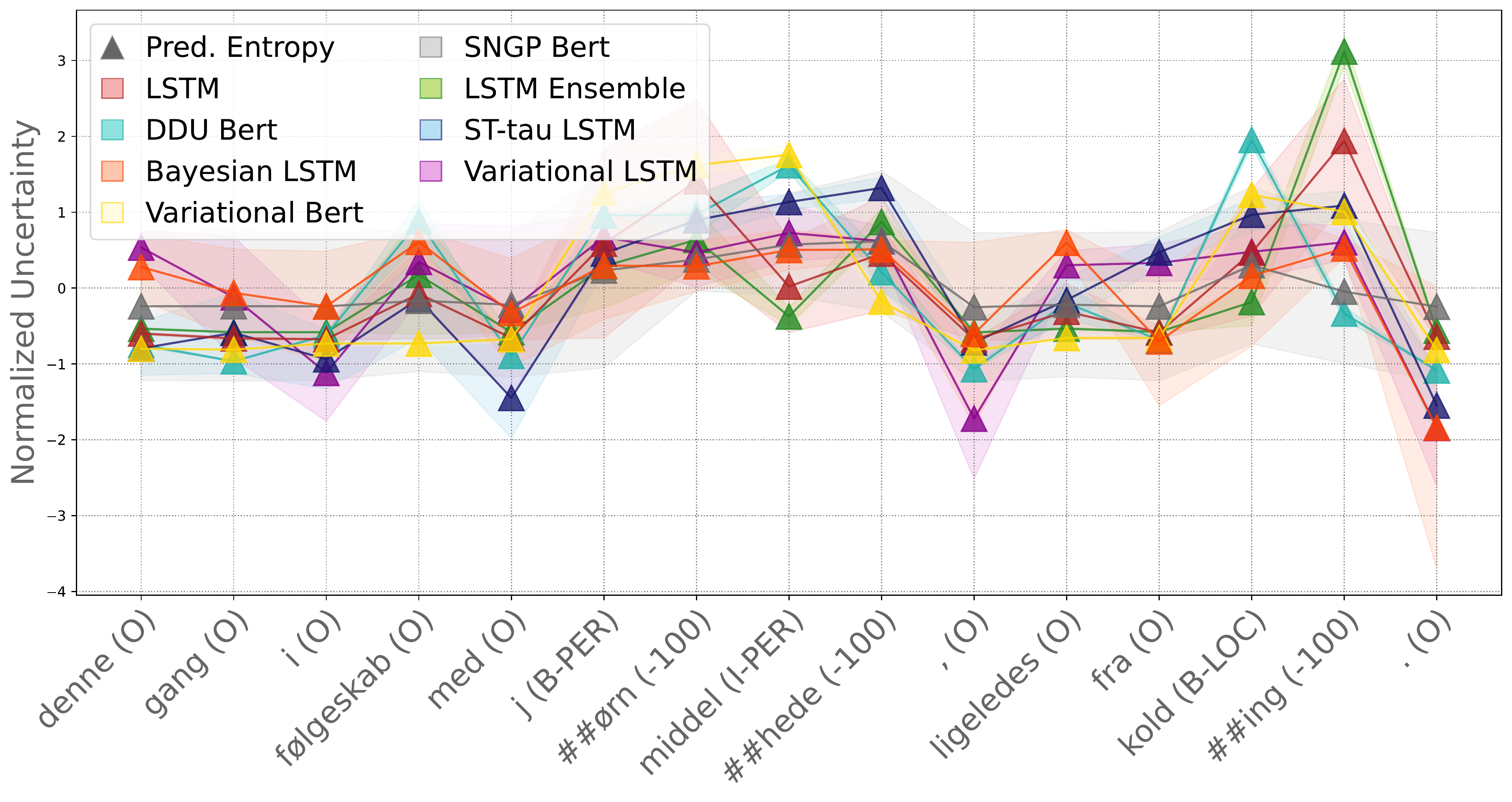}
        \subcaption{Predictive entropy over the sentence \emph{"This time in company with Jørn Middelhede, also from Kolding"}.}
    \end{subfigure}\label{subfig:qualitative-analysis-predictive-entropy-danplus}
    \hfill
    \begin{subfigure}[t]{0.99\columnwidth}
        \centering
        \includegraphics[width=\columnwidth]{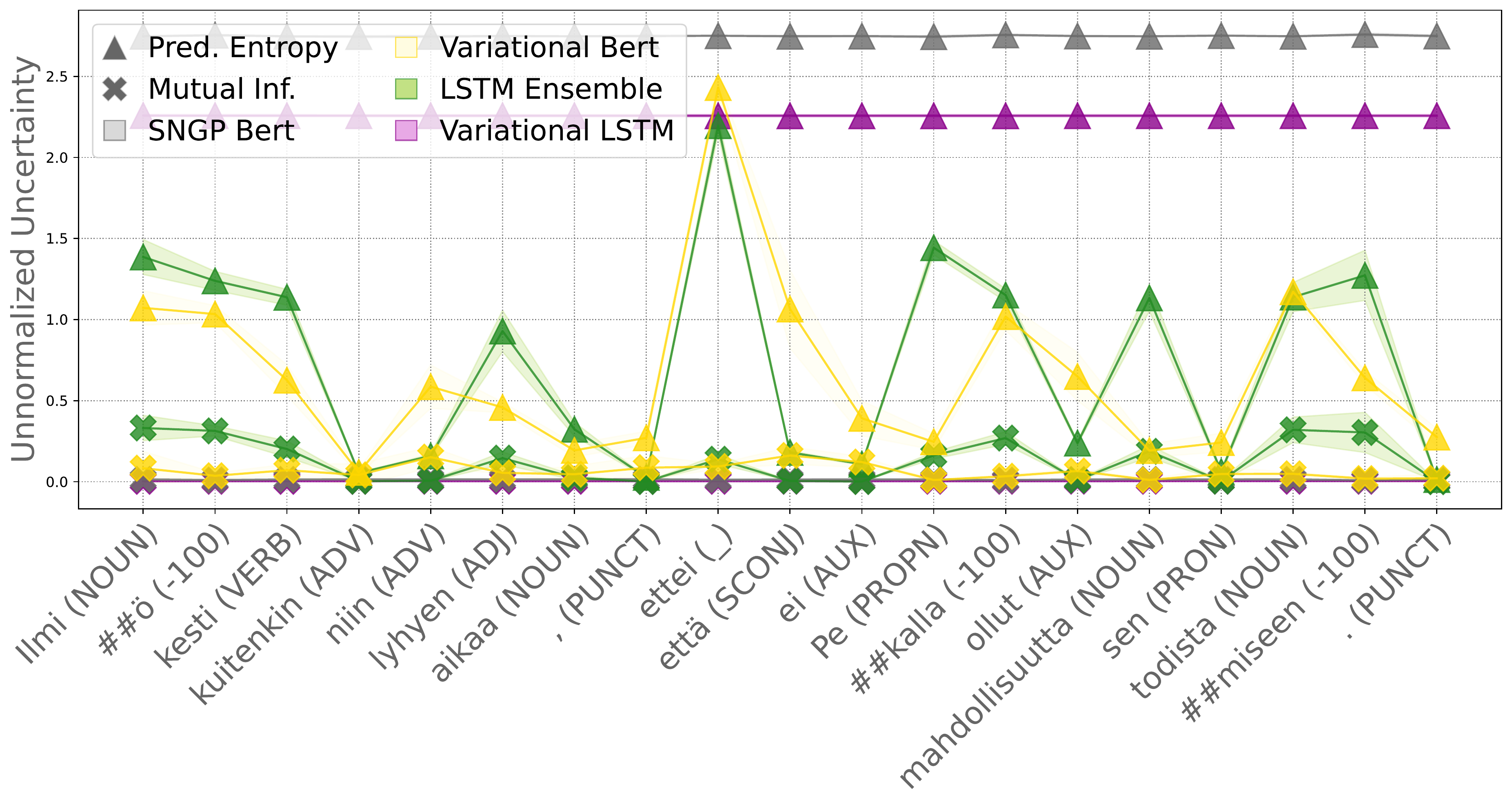}
        \subcaption{Predictive entropy and mutual information over the sentence \emph{"However, the phenomenon lasted for such a short time that Pekka did not have a chance to prove it"}.}\label{subfig:qualitative-analysis-mutual-information-finnish-ud}
    \end{subfigure}
    \caption{\textbf{Uncertainty estimates on single sequences}, for predictive entropy of different models on Danish (\cref{subfig:qualitative-analysis-predictive-entropy-danplus}) and predictive entropy and mutual information for multi-prediction models on Finnish (\cref{subfig:qualitative-analysis-mutual-information-finnish-ud}).}\label{fig:qualitative-analysis}
\end{figure*}

We investigate the development of uncertainty estimates over the course of a single sequence for different datasets, models, and uncertainty metrics. We showcase two examples in \cref{fig:qualitative-analysis}, with more examples in \cref{app:qualitative analysis}. By looking at the predictive entropy of models in \cref{subfig:qualitative-analysis-predictive-entropy-danplus}, we can observe multiple things: First of all, we can observe some degree of agreement between models and their uncertainty: Processing sub-word tokens, uncertainty seems to increase, and the total uncertainty always appears to reduce considerably on punctuation. Interestingly, the highest uncertainty seems to be produced by the DDU and Variational BERT models as well as the ensembles. 
In \cref{subfig:qualitative-analysis-mutual-information-finnish-ud}, we compare the estimates for predictive entropy and mutual information, the latter of which is supposed to only express model uncertainty. Here, uncertainty is generally low, indicating a large part of the total uncertainty might actually be of an aleatoric nature (which is the gap between triangle and cross markers of the same color, due to \cref{eq:mutual-information}). These insights indicate that while aleatoric uncertainty might be a constant factor for all models, epistemic uncertainty expectedly differs noticeably between them. We use all of these insights to discuss the choice of model next.

\section{Discussion}\label{sec:discussion}

Our experiments in \cref{sec:experiments} have uncovered interesting nuances about uncertainty estimation in Natural Language Processing. \textbf{With respect to RQ1}, we observe that fine-tuning BERTs and training LSTM ensembles on different languages produces high task scores with low calibration errors and high-quality uncertainty estimates, but only so on in-distribution data. On OOD data, uncertainty estimates from fine-tuned models do actually become less indicative of potential model loss compared to LSTM-based models. We also find that among the variety of uncertainty metrics proposed, there does not appear to be a superior metric. Differences in Kendall's $\tau$ on a token and sequence level suggest that loss and uncertainties fluctuate over the course of sequence. \textbf{Answering  RQ2}, it seems that paradoxically more training data seems to decrease the quality of uncertainty estimates on OOD data for pre-trained models. We speculate that fine-tuning models increasingly lets them forget relevant features that would produce higher uncertainty.  This might explain why for LSTM-type models, this effect seems to be smaller. Lastly, \textbf{we conclude about RQ3} that all models' total uncertainty behave somewhat similarly, potentially due to the strong influence of aleatoric uncertainty. From these insights, we conclude that the approaches using pre-trained models overall give the best trade-off between task performance, uncertainty quality and calibrations, however their failure on OOD samples opens up further directions of research. Ensembles can provide an alternative here in data-scarce settings, when the task is sufficiently learnable without the need for pre-training.

\section{Conclusion}

In this work, we explore the current options for uncertainty estimation in NLP on three different languages and tasks, focusing on the impact of available data on the quality of uncertainty scores in a potential low-resource environment. We conclude the following: Fine-tuning pre-trained models produces the best results in terms of task performance, calibration and uncertainty quality, but only on in-distribution data. On out-of-distribution data, LSTM-based models produce more reliable estimates, and could be preferred in cases pre-trained models might not be available, with LSTM-ensembles providing an especially attractive alternative. We discover that more training data seems to decrease quality of uncertainty on OOD, and show that the total uncertainty of models seems to often to be influenced by their aleatoric uncertainty.

\paragraph{Future Work} We see our work as groundwork for future research: While uncertainty estimation is a thriving subject in Computer Vision, it remains understudied in NLP. Our experiments highlight that the model behavior on language data is not well-understood and open several lines for further investigation: One such line is the development of new methods for NLP that a) produce more faithful estimates on OOD data while retaining their ID performance and b) require less training data to so, in order to be applicable in low-resource settings. Additionally, our qualitative analyses along with existing works such as \citet{xiao2021hallucination,xu2020understanding} highlighted the potential to use uncertainty to understand model behavior.

\section*{Limitations} 

Even though the experiments test a large array of models and metrics, the here shown collection is by no means exhaustive, and thus only a selection of popular models or approaches from very different families were considered. 

\paragraph{} Another glaring shortcoming is the focus on only three European languages: By comparing members of the Uralic, North Germanic and West Germanic families, we only scratch the surface when it comes to the morphological diversity of human language. Further, we only focused on languages with a latin writing systems, as well as specific text domains. This is due to resource constraints and the availability of suitable OOD test sets. We hope that follow-up works will refine our insights on a more representative sample of natural languages.

\paragraph{} Lastly, we solely focused on sequence labelling and sequence predictions tasks. \citet{van2022benchmarking} feature more sequence prediction tasks for English, however we are looking forward to similar studies on natural language generation and structured prediction tasks as well. 

\section*{Ethics Statement}

We do not foresee any immediate negative ethical consequences of our research. 

\section*{Acknowledgements}
We would like thank Mike Zhang and Joris Baan for their feedback on a draft, with a special thanks to the former for his input on the presentation of results in this work. Furthermore, we express our gratitude to Daniel Varab for verifying the translations of the Danish sentences shown in our works, and to Jenna Kanerva, Otto Tarkka, Antti Virtanen and the rest of the TurkuNLP group for the verification of the Finnish translations.
The authors also acknowledge the IT University of Copenhagen's HPC resources made available for conducting the research reported in this paper.

JF was supported by the Novo Nordisk Foundation (NNF20OC0062606 and NNF20OC0065611), the Independent Research Fund Denmark (9131- 00082B) and the Innovation Fund Denmark (0175-00014B).


\bibliography{anthology,custom}
\bibliographystyle{acl_natbib}

\newpage
\appendix

\section{Data}\label{app:data}

\subsection{Pre-processing}\label{app:pre-processing}

\paragraph{Tokenization} We use the corresponding BERT tokenizer for each language, including for LSTM-based models to ensure compatibility. For English, this corresponds to the original SentencePiece tokenizer used by \citet{devlin2019bert}, while we use the tokenizer of the Danish BERT \citep{hvingelby2020dane} and Finnish BERT \citep{virtanen2019multilingual} for those lanuages, respectively.

\paragraph{Tags for Sub-word Tokens} For named entity recognition and part-of-speech tagging, we follow \citet{jurafsky2022speech}, chapter 11.3.3 to deal with sub-word tokens: For every token that is split into sub-word tokens, we assign the tag only to the first sub-word token, and $-100$ for the rest, which ignores them for evaluation purposes.  

\subsection{Sub-sampling of Training Sets}\label{app:training-set}

Since we sub-sample some of the data splits in \cref{table:datasets}, this bears the dangers of producing unnatural samples of text. For that reason, we use this appendix to describe the sampling strategies in more detail.

\paragraph{Sub-sampling procedure} 
The procedure for subsampling text is that sequences are first placed into buckets of the same label, then into sub-buckets of the same length. Then, the sampling procedure consists of first drawing a label based on the observed label frequencies, after which the draw of sequence length, proportional to the frequency of this length inside the bucket, determines the final bucket from which a sequence is again drawn uniformly.\\
Lastly, the process for token classification involves the grouping into sequences by length at the highest level. Inside a bucket, a sequence is not drawn uniformly but with a probability according to the \emph{alignment} of the sequence's labels with the overall corpus label distribution. This alignment is calculated for each sequence by evaluating the expected log-probability of the sequence's label distribution w.r.t to the label distribution of the corpus (i.e.,\@ the cross-entropy). The scores for all same-length sequences in a bucket are then normalized into a $[0, 1]$ interval in order to enable sampling, which is similar to the two-stage procedure used in the sequence classification case.

\begin{figure*}
    \centering
    \includegraphics[width=0.95\textwidth]{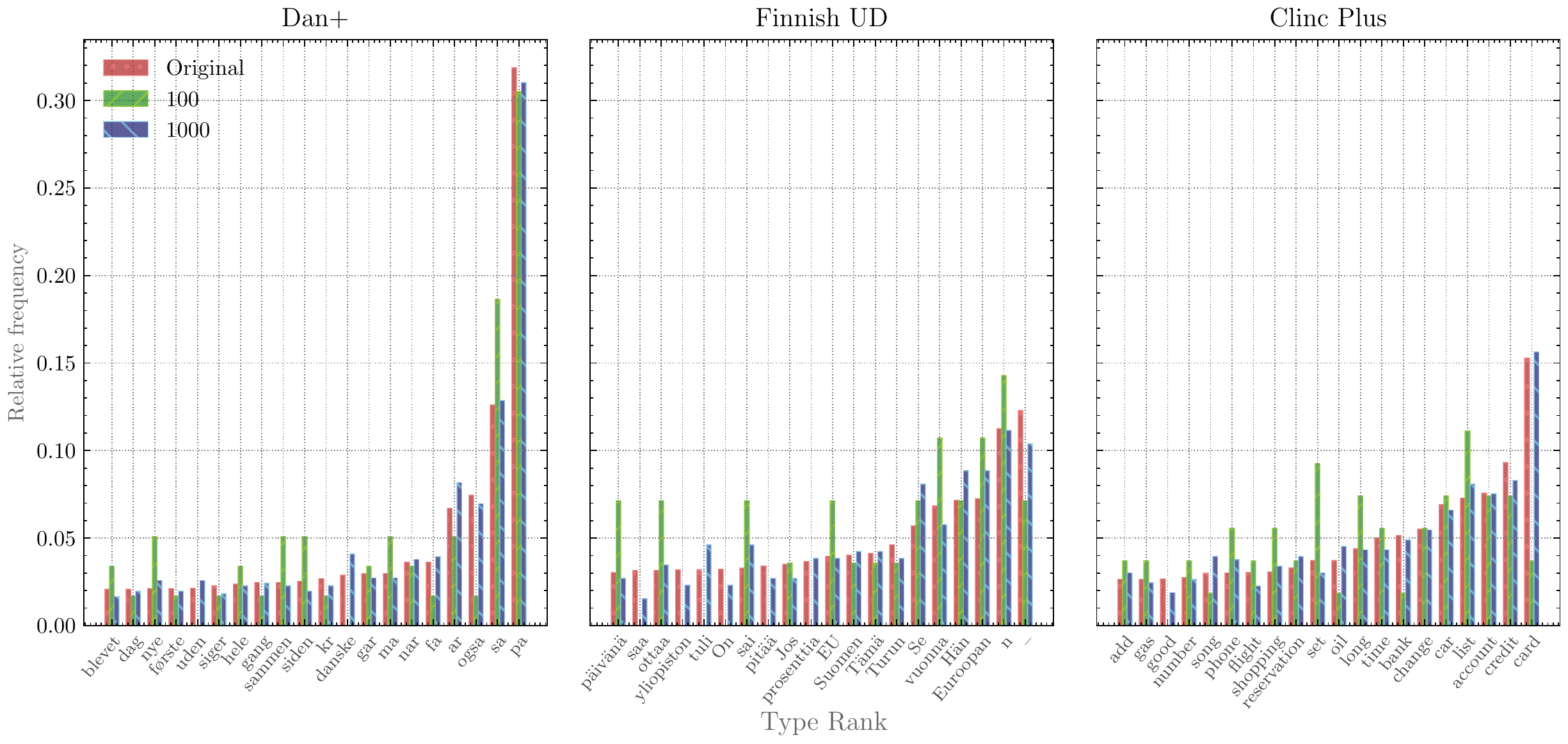}
    \caption{\textbf{Comparing the relative frequency of types in the original and sub-sampled training sets.} Shown are the top $20$ types in the original training set, compared to sub-sampled training sets of $100$ and $1000$ sequences for Dan+, Finnish UD and Clinc Plus. It is shown that while the type frequencies differ noticeably for the small dataset, already $1000$ sequences suffice to approximate the original frequencies. Numbers, stopwords and the most common punctuation were removed.}
    \label{fig:top50}
\end{figure*}

\begin{figure*}
    \centering
    \includegraphics[width=0.95\textwidth]{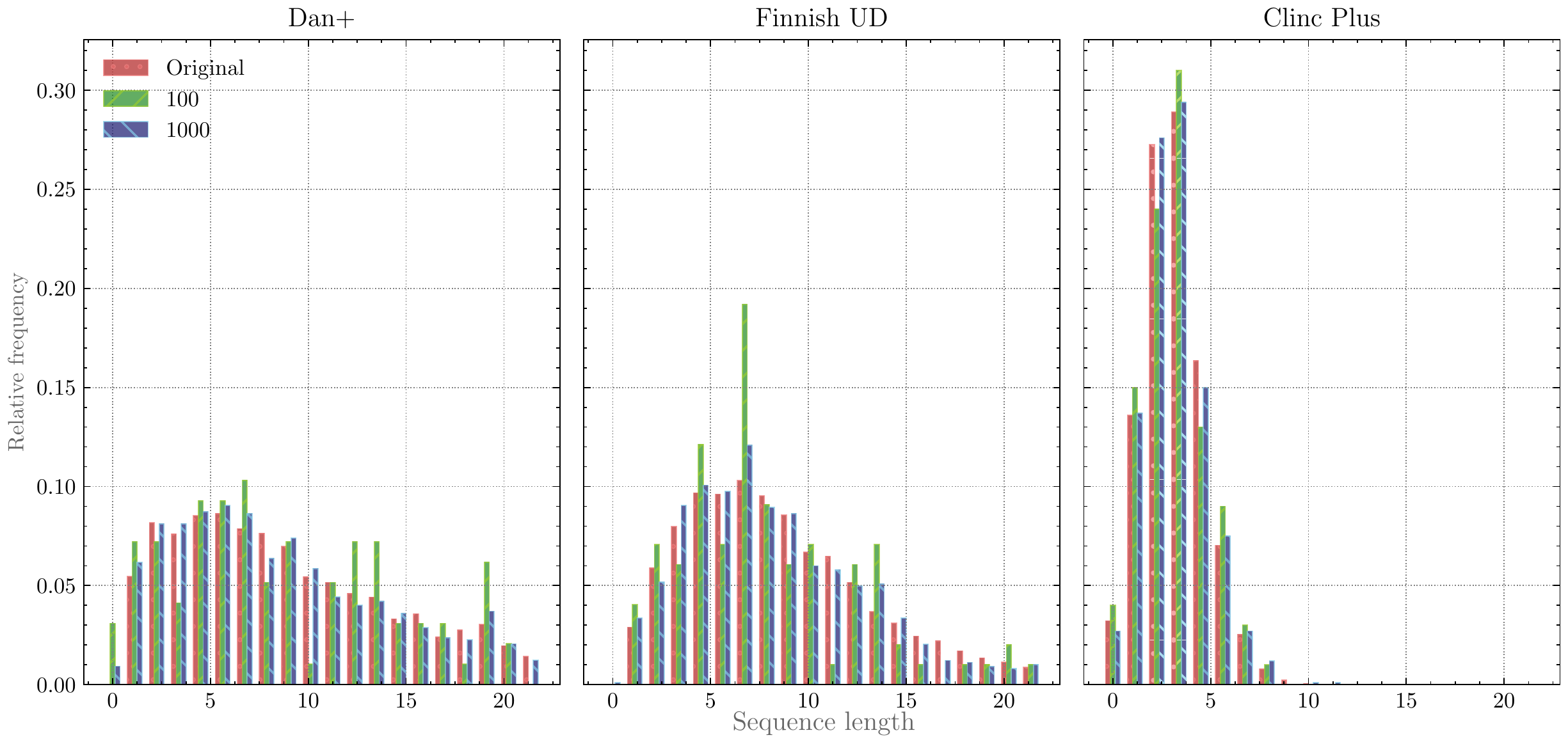}
    \caption{\textbf{Comparing the relative frequency of sequence lengths in the original and sub-sampled training sets.} Shown are sequence lengths between $0$ and $25$ in the original test, compared to OOD test sets for Dan+, Finnish UD, Clinc Plus. Not the whole distribution is shown in all cases, with many of the OOD sentences for Dan+ being very long. For Dan+ and Finnish UD, the sentence length distributions are noticeably different. For Clinc Plus, they are very similar.}
    \label{fig:sentence-lengths}
\end{figure*}

\begin{figure*}
    \centering
    \includegraphics[width=0.985\textwidth]{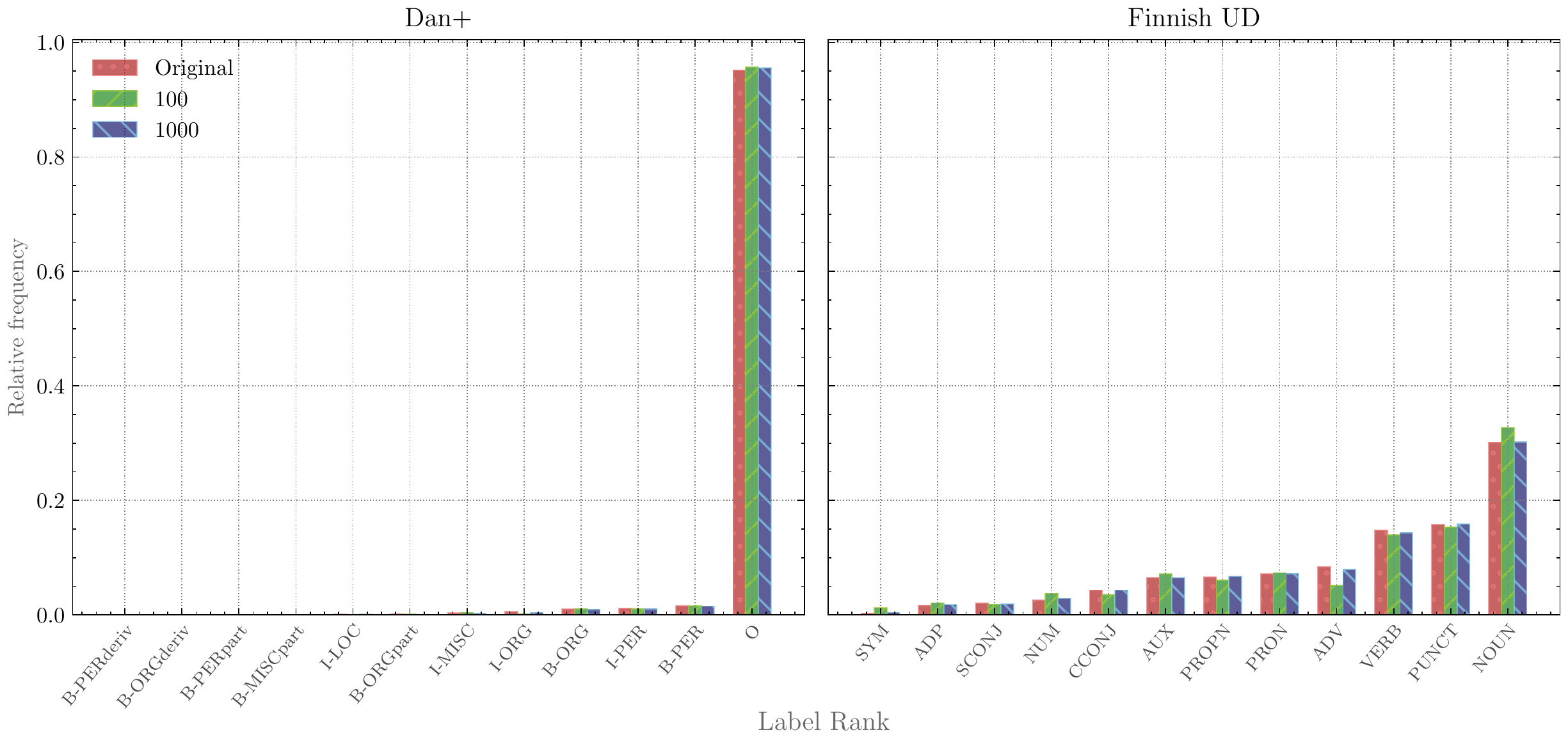} \\
    \caption{\textbf{Comparing the relative frequency of labels in the original training set, compared to sub-sampled training sets.} Shown are frequencies for $100$ and $1000$ sequences. For Danish, the most frequent label by far is the neutral label indicating that no named entity is present.}
    \label{fig:class-labels}
\end{figure*}
\begin{figure*}
    \centering
    \includegraphics[width=0.985\textwidth]{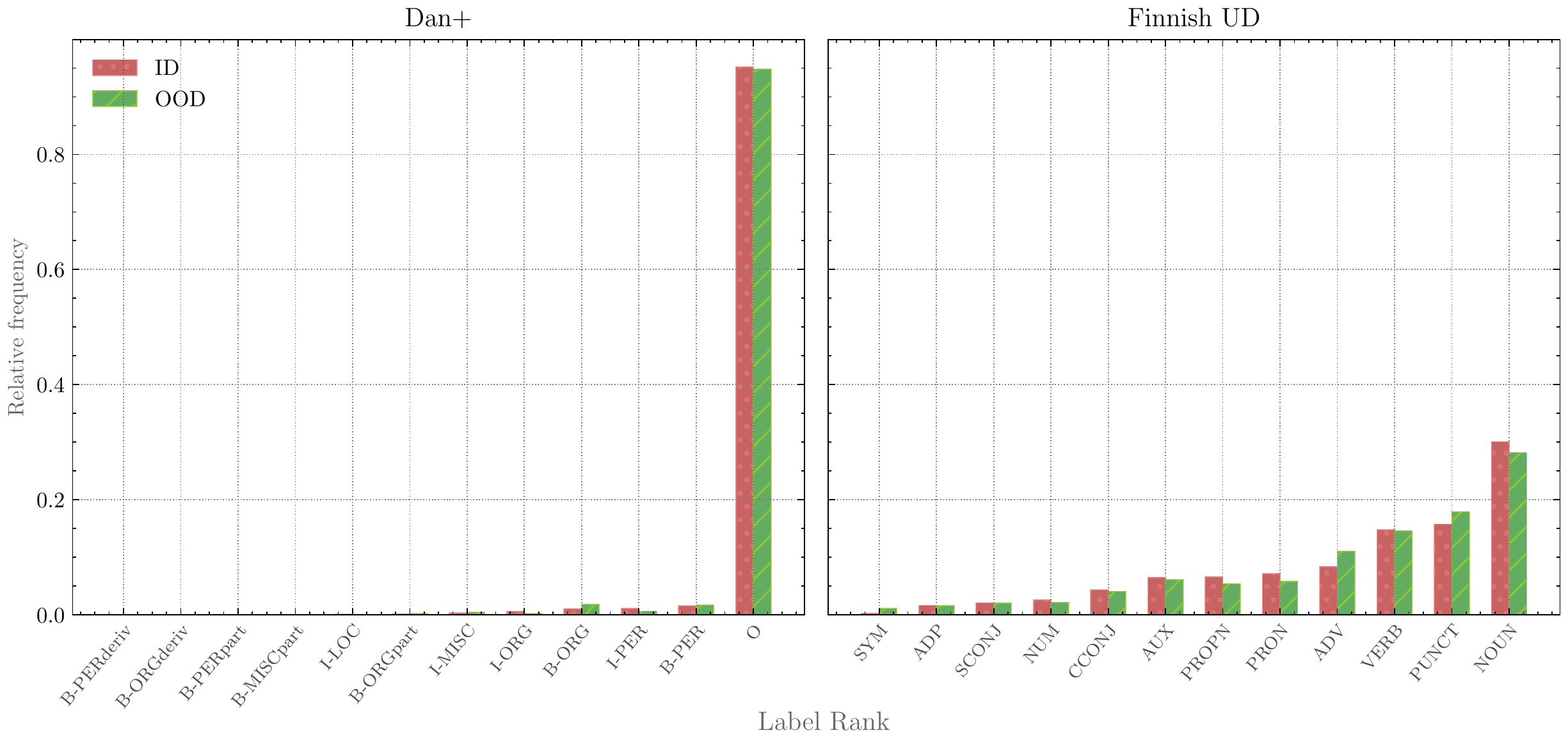}
    \caption{\textbf{Comparison of the relative class frequencies between original training set compared to the OOD test set.} The proportions stay largely the same for Danish, while different more for Finnish.}
    \label{fig:ood-class-freqs}
\end{figure*}

\paragraph{Validation of sub-sampled training sets} We take multiple steps to validate the representativeness of our sub-sampled data splits. First, we plot the distributions of the $50$ most frequent types in the original corpus in \cref{fig:top50}, where we see that distributions converge with increasing sample size. Secondly, we plot sentence length distributions in \cref{fig:sentence-lengths}, where we also see increasing alignment with sample size.\footnote{The distributions for Language Modelling are slightly distorted since we sample whole sets of sentences.} For Sequence and Token Classification tasks, we also plot the class distributions in \cref{fig:class-labels}. Lastly, we train an interpolated trigram Kneser-Ney language model \citep{jelinek1980interpolated, ney1994structuring} with uniform interpolation weights trained on the original training set using SRILM \citep{stolcke2002srilm} and sub-word tokens produced by the corresponding BERT tokenizer, sub-sample multiple splits and compare their perplexity scores to those of the original corpus in \cref{table:data-validation}. While $n$-gram perplexities of sub-sampled training sets do lie over the ones of the original data, they are still upper-bounded by the in-distribution test-set perplexities. Furthermore, this verification was not aimed to give the most precise results, as also the scoring using an $n$-gram model can be rather crude. Thus, with all these results, we conclude that our sub-sampling procedure produces sufficiently representative samples of the original data for the different tasks discussed.

\begin{table*}[ht!]
    \centering 
    \resizebox{0.85\textwidth}{!}{
        \renewcommand{\arraystretch}{2}
        \begin{tabular}{@{}lrrrrrrr@{}}
            \toprule
            Language & Train ppl.$\downarrow$ & \multicolumn{3}{c}{Sub-sampled Train ppl.$\downarrow$} & Test ppl.$\downarrow$ & OOD Test ppl.$\downarrow$\\
             &  & $n=100$ & $n=500$ & $n=1000$ & &\\
            \midrule
            English & $31.54$ & $43.97\pm2.46$ & $44.50\pm0.68$ & $44.9\pm0.4$ & $53.11$ & $ 120.32$ \\
            Danish & $112.73$ & $252.52\pm 13.25$ & $247.09\pm 3.3$ & $249.27\pm 3.15$ & $418.71$ & $524.32$ \\
            Finnish & $116.49$ & $257.67\pm 10.96$ & $257.66 \pm 4.7$ & $260.36 \pm 5.36$ & $1374.76$ & $1284.82$ \\
            \bottomrule
        \end{tabular}%
    }
    \caption{\textbf{Results of using an interpolated Kneser-Ney $n$-gram language model on selected datasets, including sub-sampled training splits and the OOD test set.} Scores of sub-sampled training sets were obtained over five different attempts.} \label{table:data-validation}
\end{table*}

\subsection{Selection of OOD Test Sets}\label{app:ood-test-set}

\begin{figure*}[tb]
    \centering
    \includegraphics[width=0.985\textwidth]{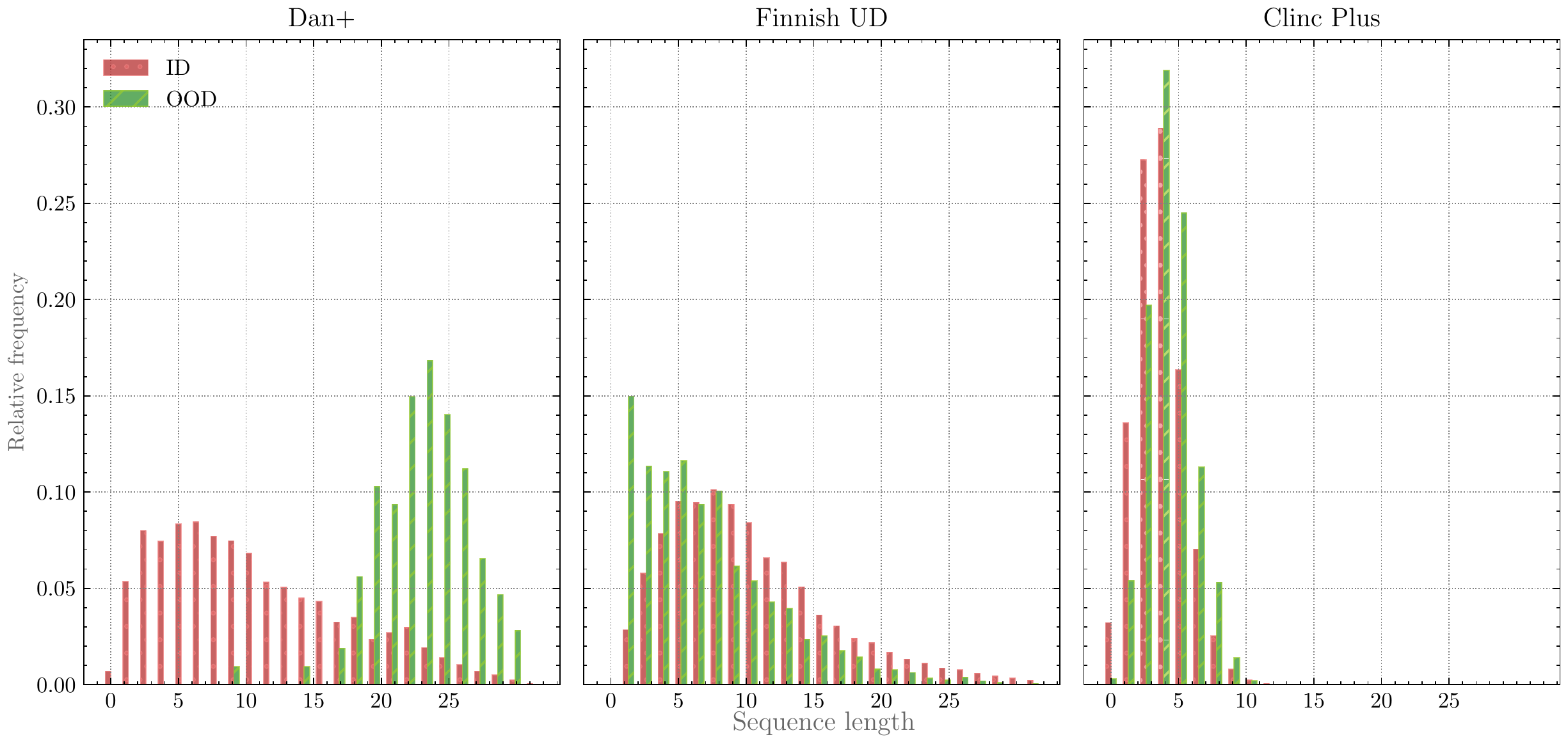}
    \caption{\textbf{Comparison of sequence length distribution between the original training set and the OOD test set.} For English, the distribution of lengths of voice assistant commands is quite similar, while the differences for Dan+ and Finnish UD are more pronounced.}
    \label{fig:ood-sequence-lengths}
\end{figure*}

\begin{figure*}[tb]
    \centering
    \includegraphics[width=0.985\textwidth]{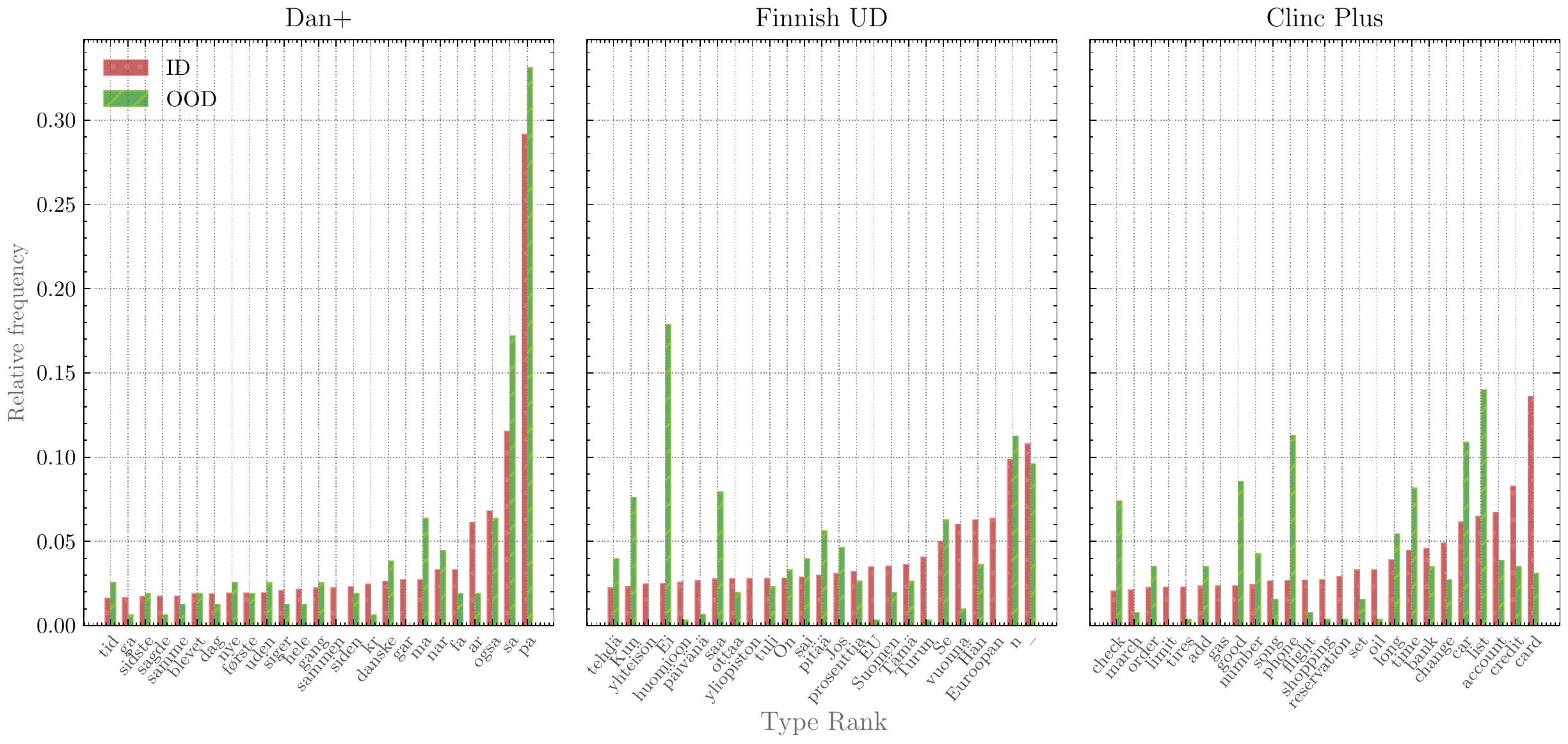}
    \caption{\textbf{Comparison of the relative frequencies of the top 25 types in the original training set compared to the OOD test set.} Even among the most frequent and therefore usually common tokens, the plots show differences between the in-distribution train and out-of-distribution test set. Numbers, stopwords and the most common punctuation were removed.}
    \label{fig:ood-type-freqs}
\end{figure*}

In this appendix section, we present additional evidence that the OOD test splits shown in \cref{table:datasets} are sufficiently different from the training data --- meaning, out-of-distribution --- to enable our chosen methodology. To that end, we re-use similar ideas as described in \cref{app:training-set}, but with the opposite goal. In \cref{fig:ood-sequence-lengths}, we plot the distribution of sequence lengths of the training set compared with the OOD test set, with the same done for the most frequent $25$ types in \cref{fig:ood-type-freqs} and class labels in \cref{fig:ood-class-freqs}. Lastly, we again use a interpolated Kneser-Ney trigram language model to compute the perplexity of the training compared to the OOD test set in \cref{table:data-validation}. In all cases, OOD $n$-gram perplexities lie much over the training or sub-sampled data perplexities. Except for Finnish, they are also widely different from the test set perplexities. In that exceptional cases, an explanation could be given by the highly agglutinative nature of Finnish, increasing the sparsity of the language despite the subword tokenization.

\section{Calibration Metrics}\label{app:calibration-metrics}

\emph{Perfect calibration} is defined as the the confidence of a neural network  corresponding to the percentage of samples with that same predicted probability actually receiving the correct label by the network. Using a predicted label $\hat{y}$ with probability $\hat{p}$, perfect calibration is defined as 

\begin{equation*}
    P(\hat{y}=y|\hat{p}=p) = p,\quad \forall p \in [0, 1]
\end{equation*}

The expected calibration error \citep{naeini2015obtaining} quantifies the difference between the confidence and the calibration on a test set by collecting predictions into $m$ bins: 

\begin{equation}\begin{aligned}\label{eq:ece}
    & \text{ECE} =  \Expect[\Big]{\hat{p}}{P\Big(\hat{y} = y \Big| \hat{p} = p\Big) - p\Big|} \\
    \approx & \sum^M_{m=1}\frac{|\mathbb{B}_m|}{N}\Big|\text{acc}(\mathbb{B}_m) - \text{conf}(\mathbb{B}_m)\Big| \\
    = & \sum^M_{m=1}\frac{1}{N}\Big|\sum_{b \in \mathbb{B}_m}\bind(\hat{y}_b = y_b) - \hat{p}_b\Big| \\
\end{aligned}\end{equation}

where $N$ is the number of data points and $\mathbb{B}_m$ denotes the $m$-th bin.\\

The problem is that ECE is only defined for binary classification and depends highly on the number of bins chosen. For the former problem, \citet{guo2017calibration} present a naive extension to multi-class classification that only considers the most likely prediction. In order to consider all classes, \citet{nixon2019measuring} introduce the static calibration error (SCE) as an extension to multi-class problems:

\begin{equation*}\label{eq:sce}
    \resizebox{\columnwidth}{!}{%
        $\displaystyle \text{SCE} = \frac{1}{K}\sum_{k=1}^K \sum_{m=1}^M \frac{N_{mk}}{N}\Big|\text{acc}(\mathbb{B}_m, k) - \text{conf}(\mathbb{B}_m, k)\Big| $
    }
\end{equation*}

Here, $N_{mk}$ denotes the number of instances of class $k$ in bin $m$, and $\text{acc}(\mathbb{B}_m, k)$, $\text{conf}(\mathbb{B}_m, k)$ the accuracies and confidences for class label $k$ in bin $m$, respectively. However, we found this error not be very informative in our case, and therefore omitted corresponding results.\\

Secondly, \citet{nixon2019measuring} introduce the adaptive calibration error (ACE), which makes sure that every bin contains the same number of predictions. They define a calibration range by the $\lfloor N/R\rfloor$-th index of sorted and thresholded predictions. Then, the error is defined as 

\begin{equation*}
    \resizebox{\columnwidth}{!}{%
    $\displaystyle \text{ACE} = \frac{1}{KR} \sum_{k=1}^K \sum_{r=1}^R \Big|\text{acc}(\mathbb{B}_m, r) - \text{conf}(\mathbb{B}_m, r)\Big| $
    }
\end{equation*}





\section{Implementation \& Experiments}

\subsection{Implementation Details}\label{app:implementation-details}

\paragraph{Resources} All models were implemented in PyTorch \citep{paszke2019pytorch}. BERT models where implemented with the help of HuggingFace's \texttt{transformers} library \citep{wolf2020transformers}. Linear algebra operations where often implemented using the \texttt{EinOps} package \citep{rogozhnikov2022einops}. The Bayesian LSTM was developed using the \texttt{Blitz} package \citep{esposito2020blitzbdl} for PyTorch and the SNGP transformer using \texttt{gpytorch} \citep{gardner2018gpytorch}. Hugginface's \texttt{datasets} \citep{lhoest2021datasets} were furthermore used for dataset creation and \texttt{codecarbon} \citep{codecarbon} for carbon emissions tracking. Weights \& Biases \citep{wandb} was used to track and manage hyperparameter searches and experiments. In general, we follow many of the experimental guidelines and suggestions laid out by \citet{ulmer2022experimental}. 

\paragraph{Models} For the DUE transformer, we used Principal Component Analysis on the latent representations for Clinc Plus to reduce the memory usage of the Gaussian Discriminant Analysis by reducing dimensionality to $64$. We initially also experimented with the usage of the DUE transformer by \citep{van2021feature}, however found that it was not trivial to create the inducing points for the Gaussian process output layer in a sequential setting. For the Variational Transformer \citep{xiao2020wat}, the authors do not specify exactly how MC Dropout is used. We use the existing dropout layers in the corresponding model, and use a number of forward passes with different dropout masks to make predictions. Since the number of classes is prohibitive for the original formulation of the SNGP transformer, we use the extension proposed by \citet{liu2022simple} in Appendix A.1 and only store one $\hat{\Sigma}^{-1}$ matrix for all classes. Furthermore, we update the matrix continuously during training and not just during the last epoch, in order to allow tracking of the predictive performance over the training time. Lastly, we also evaluate predictions using Monte-Carlo approximations instead of the mean-field approach, since this allows us to compute a wider variety of uncertainty metrics.

\paragraph{Evaluation} When computing uncertainty estimates and losses for evaluation purposes, the measurements for a number of tokens were discarded. These include the ignore token with ID $-100$, as well as the IDs corresponding to the \texttt{[EOS]}, \texttt{[SEP]}, \texttt{[CLS]} and \texttt{[PAD]} token, which might differ between tokenizers of different languages. For computing the ECE, we use $10$ bins, and $10$ value ranges for ACE. 

\paragraph{Model Comparison} We facilitate the comparison of models using the almost stochastic order test (ASO; \citeauthor{del2018optimal}, \citeyear{del2018optimal}; \citeauthor{dror2019deep}, \citeyear{dror2019deep}), as implemented by \citet{ulmer2022deep}. One distribution is stochastically dominant over the other when its cumulative distribution function is equal or larger than its counterpart at all points. In an experimental setting, that implies that a model is producing higher scores than a baseline. The ASO test measures the deviation from the stochastic order using an approach rooted in optimal transport. We use the test with a confidence level $\alpha = 0.05$ and a decision threshold of $\tau = 0.3$.

\subsection{Hyperparameters}\label{app:training-details}

We perform hyperparameter search using random sampling \citep{bergstra2012random} using hyperband scheduling \citep{li2017hyperband}\footnote{Trials might be terminated using Hyperband after $10k$ steps.} on the entire training set, even if models are trained on sub-sampled training sets later. This has the advantage of ensuring comparability between runs and eliminating suboptimal hyperparameter choices as a source of worse uncertainty estimation. We do $80$ trials for LSTM-based models, and $30$ for BERT-based models. 
Furthermore, the hyperparameters for the LSTM are identical for the LSTM ensemble (10 instances are used per ensemble). Hyperparameters were picked by best final validation loss over search trials.

\paragraph{Chosen Hyperparameters} We summarize some common hyperparameters here and show the rest in \cref{table:hyperparameter-values}. We commonly use a batch size of $32$, and sequence lengths of $35$ for LSTM-based and $128$ for BERT-based models. All LSTM-based models are trained using $2$ layers, with the exception of the vanilla LSTM and the LSTM-ensemble on Clinc Plus with $3$ layers. Their hidden size and embedding sizes are set to $650$. For all models, gradient clipping is set to $10$. For models using multiple predictions to compute uncertainty estimates, $10$ predictions are used at a time. 

\begin{table}[tb!]
    \centering 
    \resizebox{0.985\columnwidth}{!}{
        \renewcommand{\arraystretch}{2}
        \begin{tabular}{@{}lrr@{}}
            \toprule
            Name & Tuned for & Search space \\
            \midrule
            Learning rate & \makecell[tr]{LSTM, LSTM Ensemble,\\ Bayesian LSTM, ST-$\tau$ LSTM\\ Variational LSTM} & $\mathcal{U}(0.1, 0.5)$ \\ 
            Learning rate & \makecell[tr]{DDU BERT, SNGP BERT,\\ Variational BERT} & $\log \mathcal{U}(10^{-5},10^{-3})$ \\ 
            Spectral norm upper bound & DDU BERT, SNGP BERT & $\mathcal{U}(0.95, 0.99)$ \\
            Kernel amplitude & SNGP BERT & $\log \mathcal{U}(0.01, 0.5)$ \\
            $\beta$ weight decay & SNGP BERT & $\log \mathcal{U}(10^{-3}, 0.5)$\\
            Weight decay & \makecell[tr]{LSTM, LSTM Ensemble,\\ ST-$\tau$ LSTM, Variational BERT}  & $\mathcal{U}(0.1, 0.5)$ \\ 
            Layers & LSTM, LSTM Ensemble & $\{2, 3\}$  \\ 
            Dropout & \makecell[tr]{LSTM, LSTM Ensemble,\\ ST-$\tau$ LSTM, Variational BERT} & $\mathcal{U}(0.1, 0.4)$   \\ 
            Layer Dropout & Variational LSTM & $\mathcal{U}(0.1, 0.4)$   \\ 
            Time Dropout & Variational LSTM & $\mathcal{U}(0.1, 0.4)$   \\ 
            Embedding Dropout & Variational LSTM & $\mathcal{U}(0.1, 0.4)$   \\ 
            Hidden size & LSTM, LSTM Ensemble & $\{350, 500, 650\}$ \\ 
            Prior $\sigma_1$ & Bayesian LSTM & $\log \mathcal{U}(-0.8, 0.1)$ \\ 
            Prior $\sigma_2$ & Bayesian LSTM & $\log \mathcal{U}(-0.8, 0.1)$ \\ 
            Prior $\pi$ & Bayesian LSTM & $\log \mathcal{U}(0.1, 0.9)$ \\ 
            Posterior $\mu$ init & Bayesian LSTM & $\mathcal{U}(-0.6, 0.6)$ \\ 
            Posterior $\rho$ init & Bayesian LSTM & $\mathcal{U}(-8, -2)$ \\
            Init weight & LSTM & $\mathcal{U}(0.1, 0.4)$ \\
            Number of centroids & ST-$\tau$ LSTM & $\{5, 10, 20, 30, 40\}$  \\
            \bottomrule
        \end{tabular}%
    }\caption{\textbf{List of searched hyperparameters.} LSTM Ensemble hyperparameters are not searched, but simply copied from the found LSTM hyperparameters.} \label{table:hyperparameter-ranges}
\end{table}

\begin{table}[!tb]
    \centering 
    \resizebox{0.975\columnwidth}{!}{
        \renewcommand{\arraystretch}{2}
        \begin{tabular}{@{}ll|rrr@{}}
            \toprule
            Model & Hyperparameter & English & Danish & Finnish \\
            \midrule
            LSTM & Weight decay & $ 0.001337$ & $0.001357$ & $0.001204$ \\
            & Learning rate & $0.4712$ & $0.4931$ & $0.2205$ \\
            & Init. weight & $ 0.283$ & $0.5848$ & $0.5848$ \\
            & Dropout & $0.3379$ & $0.2230$ & $0.1392$ \\
            Variational LSTM & Weight decay & -- & $10^{-7}$ & $0.01953$ \\
            & Learning rate & -- & $0.3031$ & $0.7817$ \\
            & Init. weight & -- & $0.1097$ & $0.5848$ \\
            & Embedding Dropout & -- & $0.1207$ & $0.3566$ \\
            & Layer Dropout & -- & $0.1594$ & $0.3923$ \\
            & Time Dropout & -- & $0.1281$ & $0.1646$ \\
            Bayesian LSTM & Weight decay & $0.001341$ & $0.003016$ & $0.03229$ \\
            & Learning rate & $0.1704$ & $0.1114$ & $0.1549$ \\
            & Dropout & $0.3410$ & $0.3868$ & $0.331$ \\
            & Prior $\sigma_1$ & $0.9851$ & $0.7664$ & $0.3246$ \\ 
            & Prior $\sigma_2$ & $0.5302$ & $0.851$ & $0.5601$ \\ 
            & Prior $\pi$ & $1$ & $1$ & $0.1189$ \\ 
            & Posterior $\mu$ init & $-0.005537$ & $-0.0425$ & $0.4834$\\ 
            & Posterior $\rho$ init & $-7$ & $ -6$ & $0.1124$ \\
            ST-$\tau$ LSTM & Weight decay & -- & $0.001189$ & $0.0007857$ \\
            & Learning rate & -- & $0.01979$ & $0.3601$ \\
            & Dropout & -- & $0.1867$ & $0.1737$ \\
            & Num. centroids & -- & $5$ & $30$ \\
            DDU Bert & Learning Rate & $0.003077$ & $0.00006168$ & $0.001825$ \\
             & Spectral norm upper bound & $0.9753$ & $0.9211$ & $0.941$ \\
             & Weight decay & $0.003$ & $0.1868$ & $0.09439$ \\
            Variational BERT & Learning Rate & $0.0002981$ & $0.00009742$ & $0.00003483$ \\
             & Weight decay & $0.01591$ & $0.02731$ & $0.09927$ \\
             & Dropout & $0.2382$ & $0.4362$ & $0.4364$ \\
            SNGP Bert & Learning Rate & -- & $0.0002332$ & $0.0002919$ \\
             & Spectral norm upper bound & -- & $0.99$ & $0.96$ \\
             & Beta Weight decay & -- & $0.001619$ & $0.002438$ \\
             & Beta length scale & -- & $2.467$ & $2.254$ \\
             & Kernel amplitude & -- & $0.3708$ & $0.2466$ \\
            \bottomrule
        \end{tabular}%
    }
    \caption{\textbf{List of used model hyperparameters by dataset}.} \label{table:hyperparameter-values}
\end{table}

\subsection{Optimization} \label{app:transformer-fine-tuning}

To make sure that all models are trained for the same number of steps regardless of the the size of (sub-sampled) training set, we set the training duration to the number of steps corresponding to a number of epochs using the original training set size, and name it \emph{epoch-equivalents} in the following. Due to the imbalance of classes in Finnish UD and Dan+, all models were trained using loss-weights that are inverse to the frequency of a label in the dataset.

\paragraph{Optimization of LSTMs} We adopt different optimization schemes for transformer and LSTM-based models. For LSTMs, we choose stochastic gradient descent with a decaying learning rate schedule, decaying by $0.8695$ after the equivalent of $14$ epochs for every following epoch-equivalent for $55$ epoch-equivalents in total. This corresponds to the setup in \citet{gal2016theoretically}, modified from the setup in \citet{zaremba2014recurrent}.

\paragraph{Optimization of BERTs} We fine-tune BERT models using the shorter duration of $20$ epoch-equivalents, corresponding to the NLP experiments in \citet{liu2022simple}. Adam \citep{kingma2015adam} is used for optimization with default parameters $\beta_1 = 0.9$ and $\beta_2 = 0.999$ alongside a triangular learning rate, using the first $10 \%$ of the training duration as warm-up.


\subsection{Convergence on Clinc Plus}\label{app:convergence-clinc-plus}

Here, we briefly address the models missing from the English Clinc Plus experiments. For the ST-$\tau$ and Variational LSTM, we could not identify clear reasons on why models did not converge. Even after extensive hyperparameter searches and manual fine-tuning of hyperparameters (including different learning rate schedules and optimizers), we did not find a combination of options that resulted in convergence. We also observed strange behavior for the Bayesian LSTM, which, after reaching a validation accuracy of $0.5$, would suddenly return to its initial training performance. This could potentially be explained by the model accidentally escaping a low-loss basin due to a learning rate that is still too high, and thus we changed the model to only be trained for $18$ epoch-equivalents and initiate the learning rate decay after seven epoch-equivalents. The puzzling fact is that SNGP BERT did not converge on Clinc Plus, since the authors successfully used the dataset in their own work \citep{liu2022simple}. We put forth the following explanations: First of all, we observed the model to generally possess a high variance, as demonstrated by the standard deviation on the Danish and Finnish data. Secondly, we make at least two changes to their implementation: Instead of using the mean-field approximation to the predictive distribution, we use the Monte Carlo approximation in order to compute metrics such as mutual information. Also, we update the covariance matrix $\hat{\Sigma}$ over the whole training time in order to track the predictive performance for our experiments, and not just during the last epoch.

\subsection{Environmental Impact}\label{sec:environmental-impact}

The carbon efficiency was estimated to be 0.61 kgCO$_2$eq/kWh. 
735 hours of computation were performed on a Tesla V100 GPU. This includes hyperparameter search, failed runs, debugging, and discarded runs. As a rough upper bound, we estimate the compute time for a single replication of all experiments to take around 73 hours.\footnote{Note that this number could be reduced further by using better hardware acceleration, larger batch sizes, and slightly reducing the training duration for some models. Most importantly, this number also includes compute used for hyperparameter search.} To lessen the environmental impact, all models and model predictions are published in the open-source repository. Total emissions are estimated to be 52.45 kgCO$_2$eq.
We use direct air capture by climeworks to offset the emissions \citep{climeworks}. Estimations were conducted using the \texttt{codecarbon} package \citep{codecarbon}, a joint effort from authors of \citet{lacoste2019quantifying} and \citet{lottick2019nergy}.


\section{Additional Results}\label{app:additional-results}

This section contains additional experimental results and plots that could not be added to the paper due to spatial constraints. We roughly follow the structure of \cref{sec:experiments}.

\subsection{Additional Scatter Plots}\label{app:additional-scatters}

This section provides some additional scatter plots. For all plots presented here as well as \cref{fig:scatter-plot-danplus-kendalls-tau-token}, some slight jitter sampled from $\mathcal{N}(0, 0.01)$ was added to x and y-coordinates to increase readability of overlapping points.

\begin{figure*}[htb!]
    \centering
    \centering
    \begin{subfigure}{\textwidth}
        \centering
        \includegraphics[width=\columnwidth]{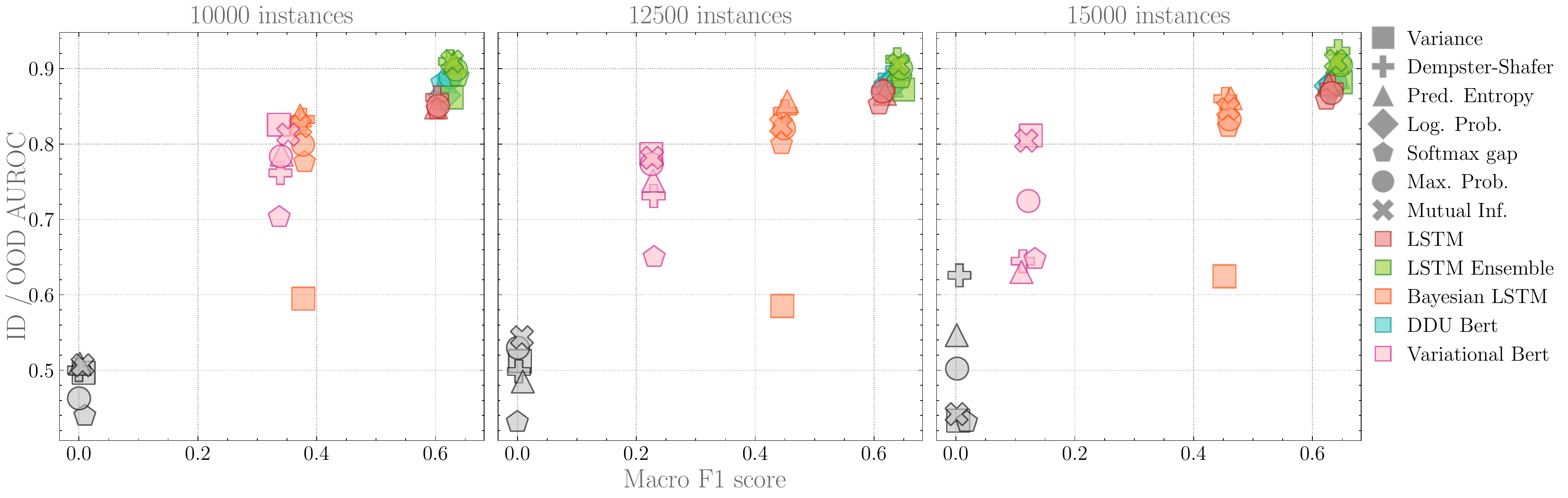}
        \subcaption{Scatter plot for the Clinc Plus dataset.}
        \label{subfig:clinc-plus-scatter-auroc}
    \end{subfigure}
    \begin{subfigure}{\textwidth}
        \centering
        \includegraphics[width=\columnwidth]{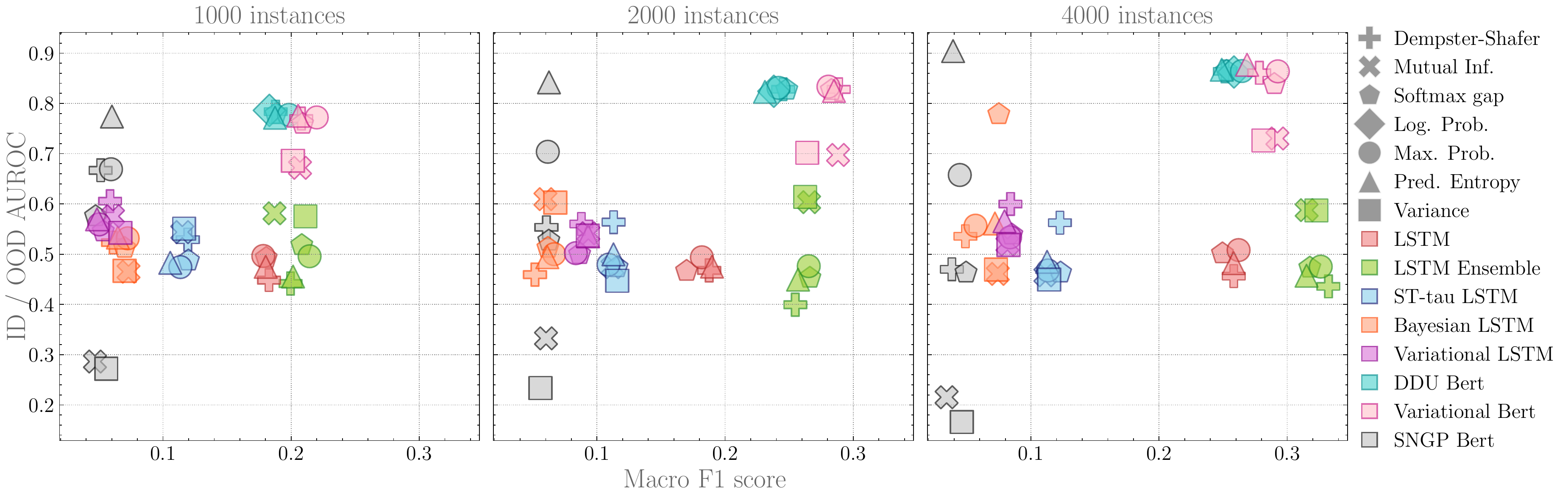}
        \subcaption{Scatter plot for the Dan+ dataset.}
        \label{subfig:danplus-scatter-auroc}
    \end{subfigure}
    \begin{subfigure}{\textwidth}
        \centering
        \includegraphics[width=\columnwidth]{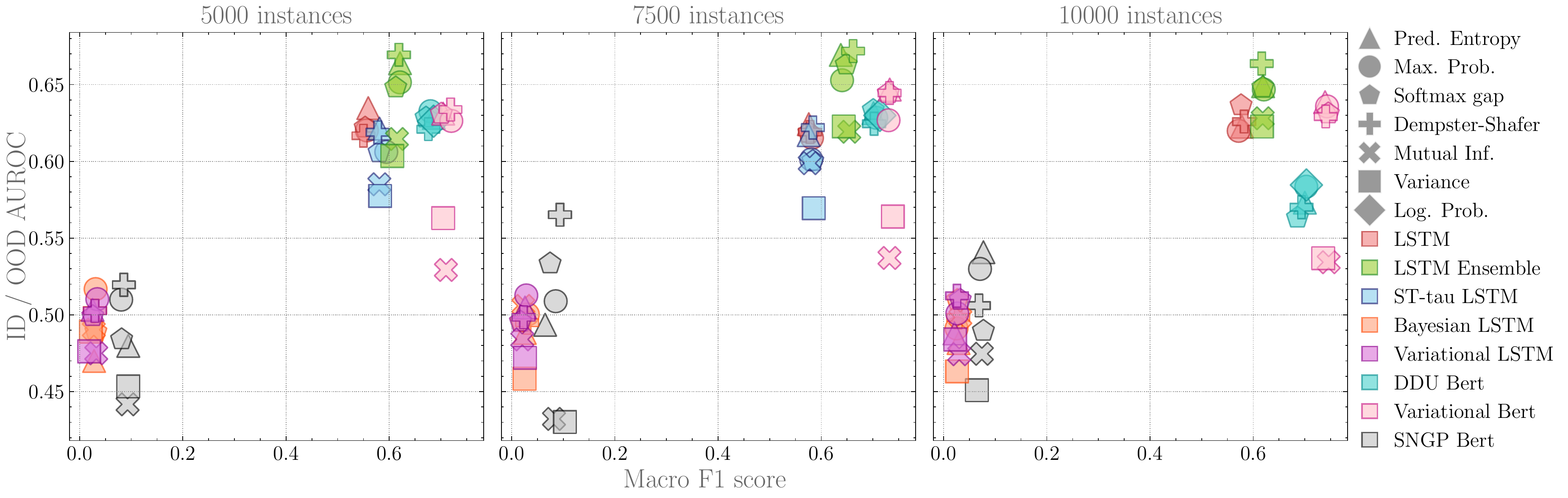}
        \subcaption{Scatter plot for the Finnish UD dataset.}
        \label{subfig:finnish-ud-scatter-auroc}
    \end{subfigure}
    \caption{\textbf{Scatter plot showing the difference between model performance (measured by macro $F_1$) and the quality of uncertainty estimates using AUROC}. Shown are different models and uncertainty metrics and several training set sizes on the used datasets.}\label{fig:scatter-plot-auroc}
\end{figure*}

\begin{figure*}[htb!]
    \centering
    \begin{subfigure}{\textwidth}
        \centering
        \includegraphics[width=\columnwidth]{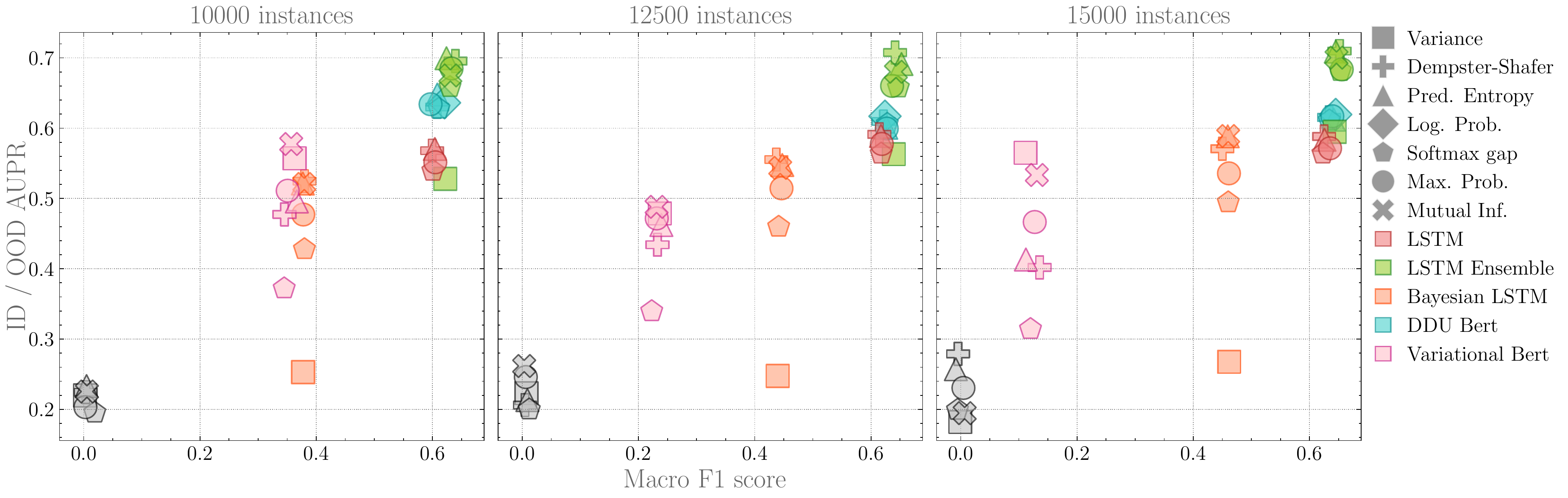}
        \subcaption{Scatter plot for the Clinc Plus dataset.}
        \label{subfig:clinc-plus-scatter-aupr}
    \end{subfigure}
    \begin{subfigure}{\textwidth}
        \centering
        \includegraphics[width=\columnwidth]{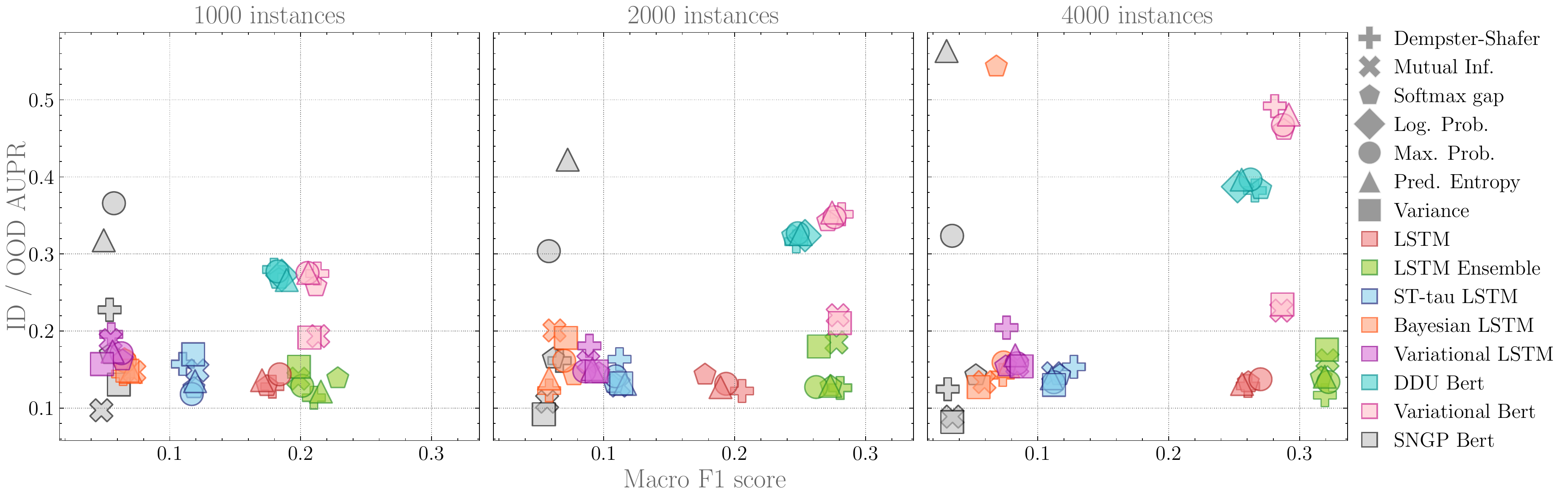}
        \subcaption{Scatter plot for the Dan+ dataset.}
        \label{subfig:danplus-scatter-aupr}
    \end{subfigure}
    \begin{subfigure}{\textwidth}
        \centering
        \includegraphics[width=\columnwidth]{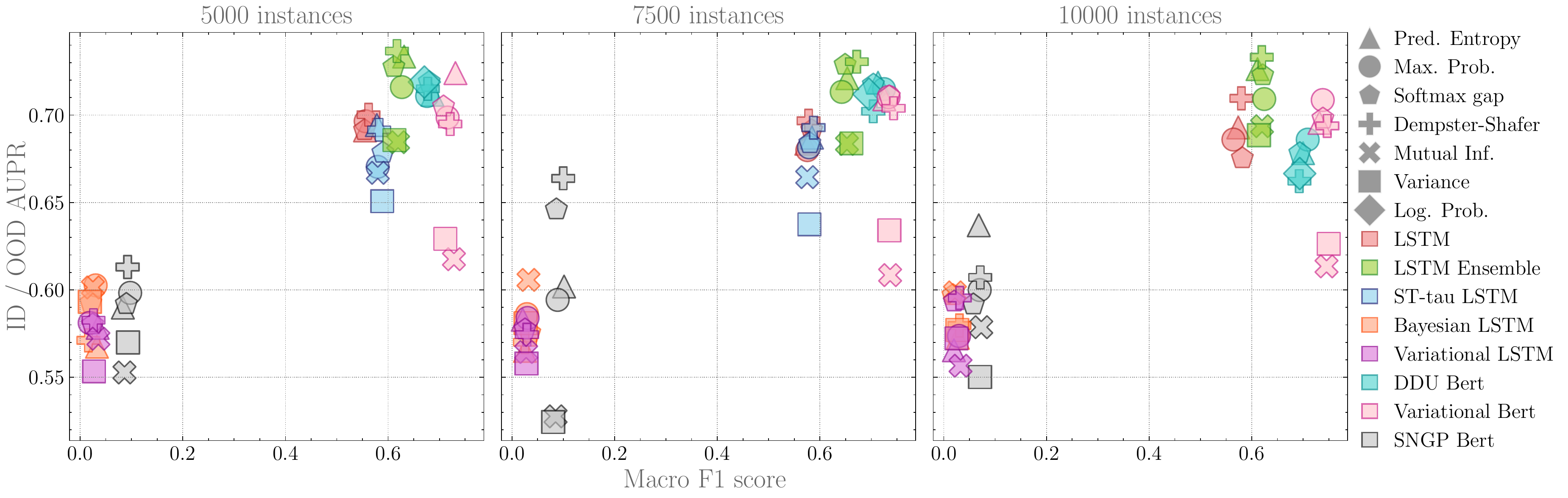}
        \subcaption{Scatter plot for the Finnish UD dataset.}
        \label{subfig:finnish-ud-scatter-aupr}
    \end{subfigure}
    \caption{\textbf{Scatter plot showing the difference between model performance (measured by macro $F_1$) and the quality of uncertainty estimates using AUPR}. Shown are different models and uncertainty metrics and several training set sizes on the used datasets.}\label{fig:scatter-plot-aupr}
\end{figure*}

\paragraph{Clinc Plus} In \cref{subfig:clinc-plus-scatter-auroc,subfig:clinc-plus-scatter-aupr}, we can see that the Variational Bert model actually \emph{degrades} in performance as the more training data is added, both on a task and uncertainty dimensions, while other models stay relatively constant. The same trend can be detected using the sequence-level Kendall's $\tau$ for Clinc Plus. We suspect that the smallest training size of $10k$ examples does already provide enough data for models to converge to similar solutions even after adding more data, and that the Variational Bert alone might be prone to overfitting in this case. 

\begin{figure*}[htb!]
    \centering
    \begin{subfigure}{\textwidth}
        \centering
        \includegraphics[width=\columnwidth]{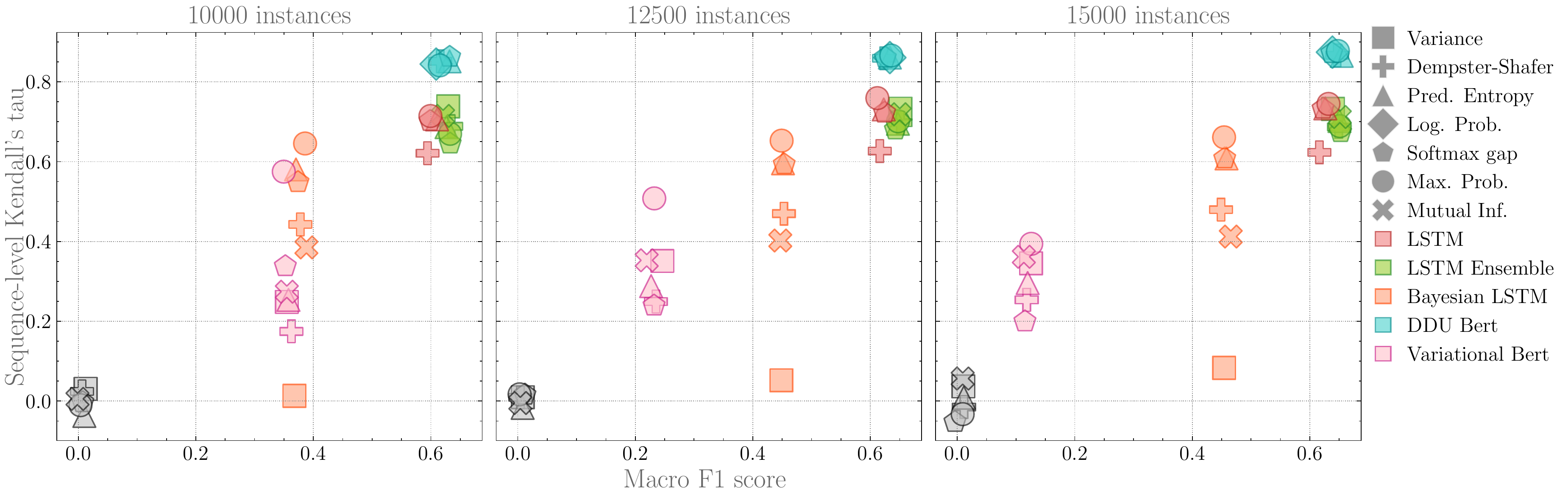}
        \subcaption{Scatter plot for the Clinc Plus dataset.}
        \label{subfig:clinc-plus-scatter-kendalls-tau-seq}
    \end{subfigure}\\
    \begin{subfigure}{\textwidth}
        \centering
        \includegraphics[width=\columnwidth]{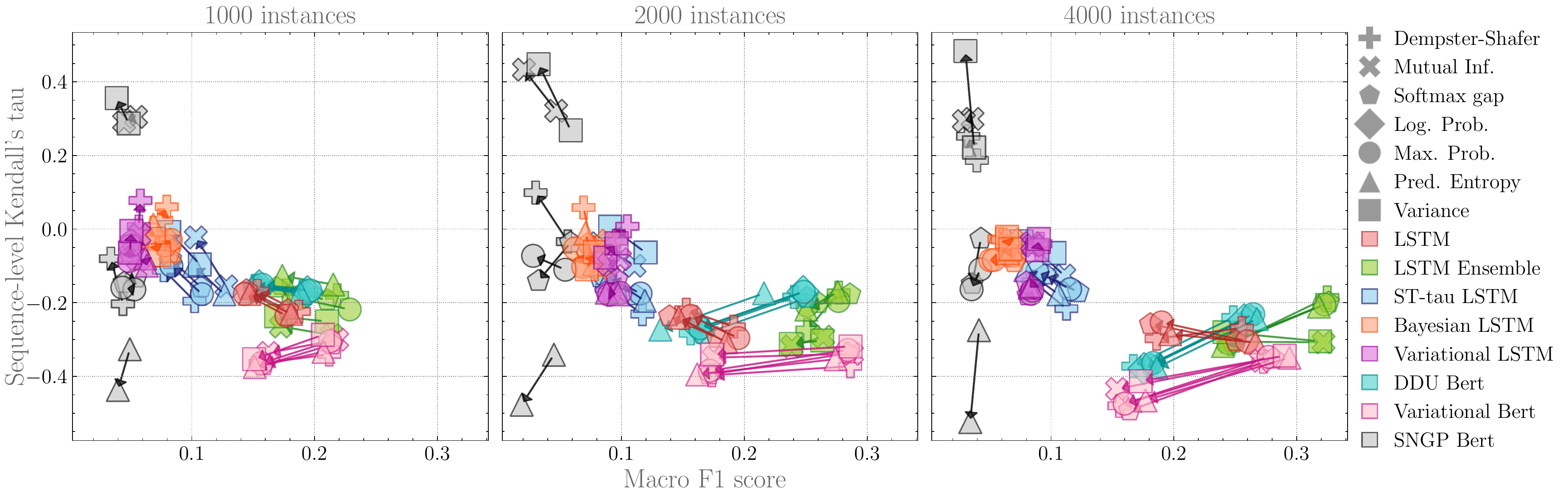}
        \subcaption{Scatter plot for the Dan+ dataset.}
        \label{subfig:dan+-scatter-kendalls-tau-seq}
    \end{subfigure}\\
    \begin{subfigure}{\textwidth}
        \centering
        \includegraphics[width=\columnwidth]{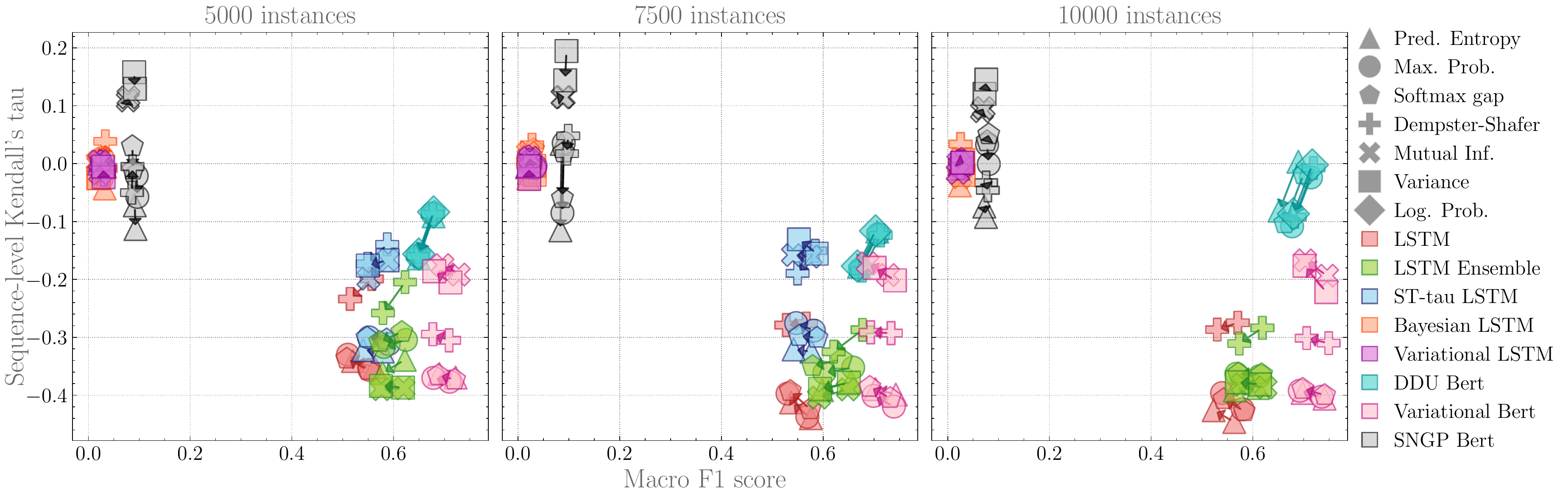}
        \subcaption{Scatter plot for the Finnish UD dataset.}
        \label{subfig:finnish-ud-scatter-kendalls-tau-seq}
    \end{subfigure}
    \caption{\textbf{Scatter plot showing the difference between model performance (measured by macro $F_1$) and the quality of uncertainty estimates on a sequence-level (measured by Kendall's $\tau$)}. Results are shown for different models and uncertainty metrics and several training set sizes on the Finnish UD and Clinc Plus dataset. Arrows indicate changes between the in-distribution and out-of-distribution test set.}\label{fig:scatter-plot-kendalls-tau-seq}
\end{figure*}

\paragraph{Dan+} Results for the Danish dataset are shown in \cref{subfig:danplus-scatter-auroc,subfig:danplus-scatter-aupr}. It is apparent that LSTM--based models stay mostly constant in their predictive performance, with the largest gains observed by the LSTM ensemble. We can also observe the DDU and Variational BERT to increase both in task performance and uncertainty quality with increasing training data. Interestingly, we can see for the SNGP BERT that uncertainty estimates become more indicative of OOD with more training samples, but mostly only using predictive entropy and the maximum probability score. This might indicate that in these cases, the model actually achieves the desired distance-awareness posed by \citet{liu2022simple}. In \cref{subfig:dan+-scatter-kendalls-tau-seq}, we can see a similar behavior of the SNGP-BERT and its metrics w.r.t. to the sequence-level correlation. Also, we see that the other BERT models and LSTM-Ensemble actually loose in uncertainty quality as more data is added.

\begin{figure*}[htb!]
    \centering
    \centering
    \includegraphics[width=2\columnwidth]{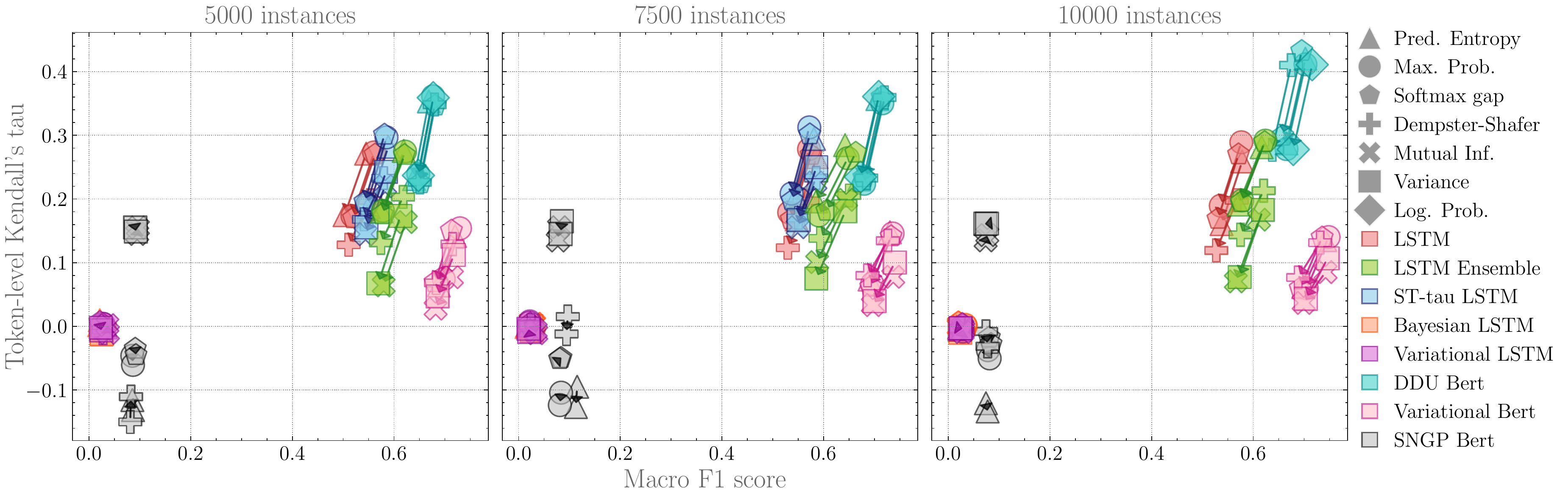}
    \caption{\textbf{Scatter plot showing the difference between model performance (measured by macro $F_1$) and the quality of uncertainty estimates on a token-level (measured by Kendall's $\tau$)}. Results are shown for different models and uncertainty metrics and several training set sizes on the Finnish UD dataset. Arrows indicate changes between the in-distribution and out-of-distribution test set.}\label{fig:finnish-ud-scatter-plot-kendalls-tau-token}
\end{figure*}

\paragraph{Finnish UD} In \cref{subfig:finnish-ud-scatter-auroc,subfig:finnish-ud-scatter-aupr}, we see that the AUROC and AUPR scores of differnet models and metrics stay largely constant across dataset sizes, which could be explained with the larger amount of training data supplied compared to Dan+.  On the token-level correlation between uncertainty and loss in \cref{fig:finnish-ud-scatter-plot-kendalls-tau-token}, we see the DDU BERT profiting most from more data. On a sequence-level, as depicted in \cref{subfig:finnish-ud-scatter-kendalls-tau-seq}, the correlation appers mostly static across training set sizes, with only small gaps between in-distribution and out-of-distribution data. 

Overall, it seems that the range of dataset sizes for Dan+ show the most critical differences between models, while for the dataset sizes used for Finnish UD and Clinc Plus, enough data seems to be supplied for changes to be more miniscule. This result is particularly relevant for low-resource setting, although the dependency on the task can not be disentangled from these results. 

\subsection{Additional Uncertainty over Training Plots}\label{app:more-uncertainty-over-training}

\begin{figure}[htb!]
    \centering
    \begin{subfigure}{\columnwidth}
        \centering
        \includegraphics[width=\columnwidth]{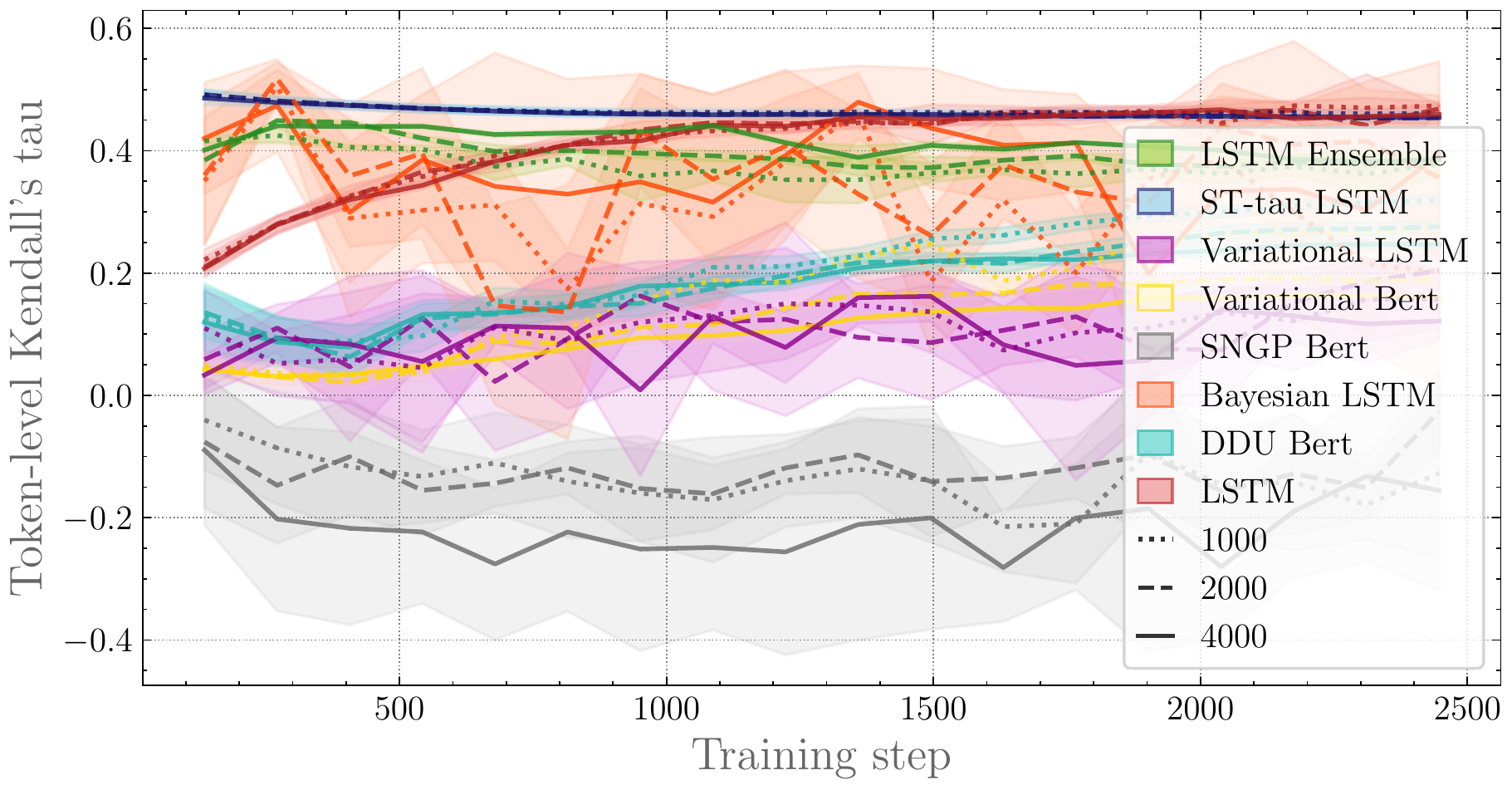}
        \subcaption{Development of token-level Kendall's $\tau$ for Dan+.}
    \end{subfigure}\label{subfig:development-models-finnish-ud-predictive-entropy-token}
    \par\bigskip 
    \begin{subfigure}{\columnwidth}
        \centering
        \includegraphics[width=\columnwidth]{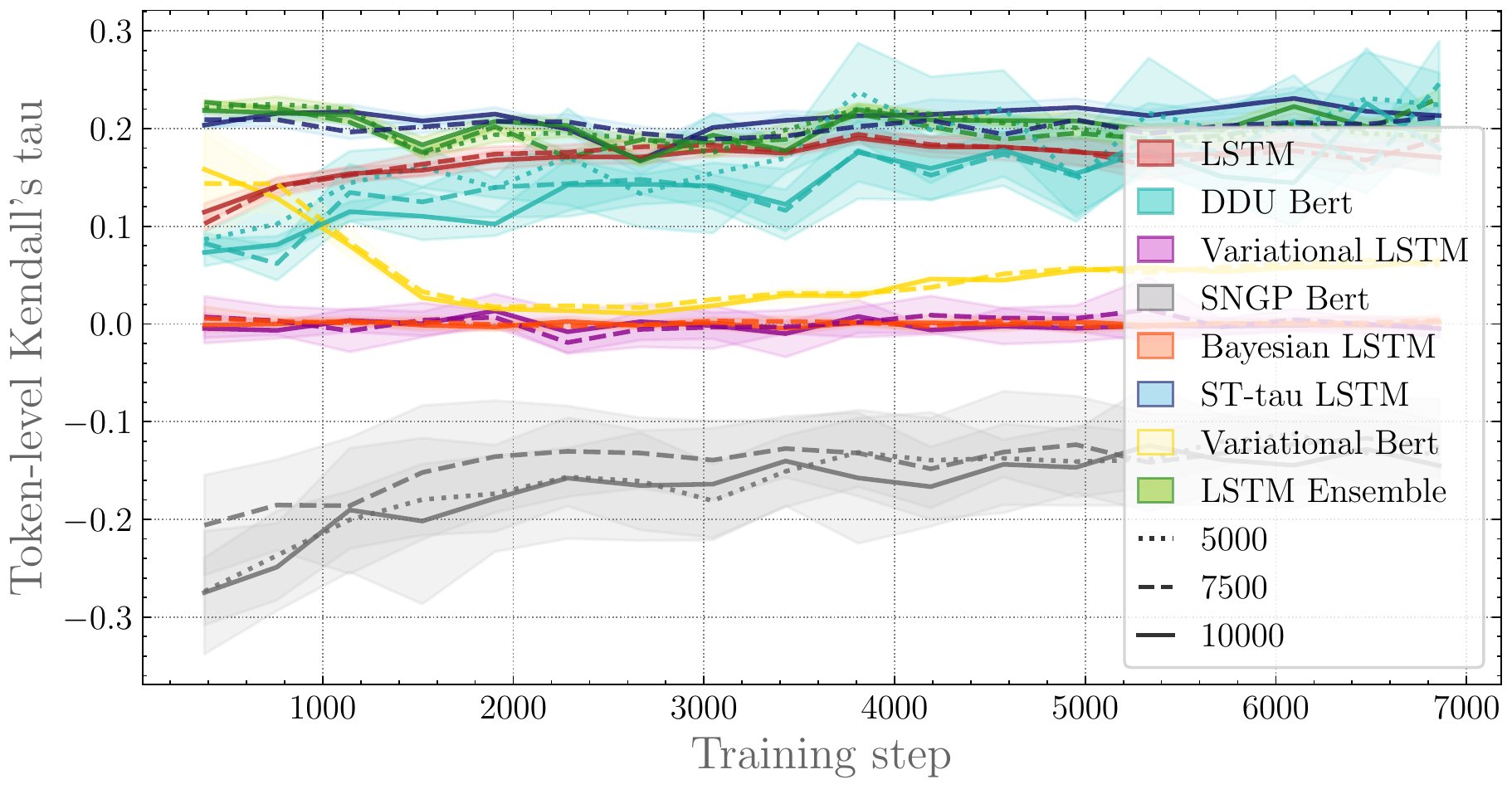}
        \subcaption{Development of token-level Kendall's $\tau$ for Finnish UD.}\label{subfig:development-models-finnish-ud-predictive-entropy-seq}
    \end{subfigure}
    \caption{\textbf{Development of correlation between token-level predictive entropy and loss on the Dan+ and Finnish UD OOD test set over the training time}. Data is shown for several model types and using differently-sized training sets. Colored areas indicate the standard deviation over five runs.}\label{fig:development-models-predictive-entropy-token-level}
\end{figure}

\begin{figure}[htb!]
    \centering
    \begin{subfigure}{\columnwidth}
        \centering
        \includegraphics[width=\columnwidth]{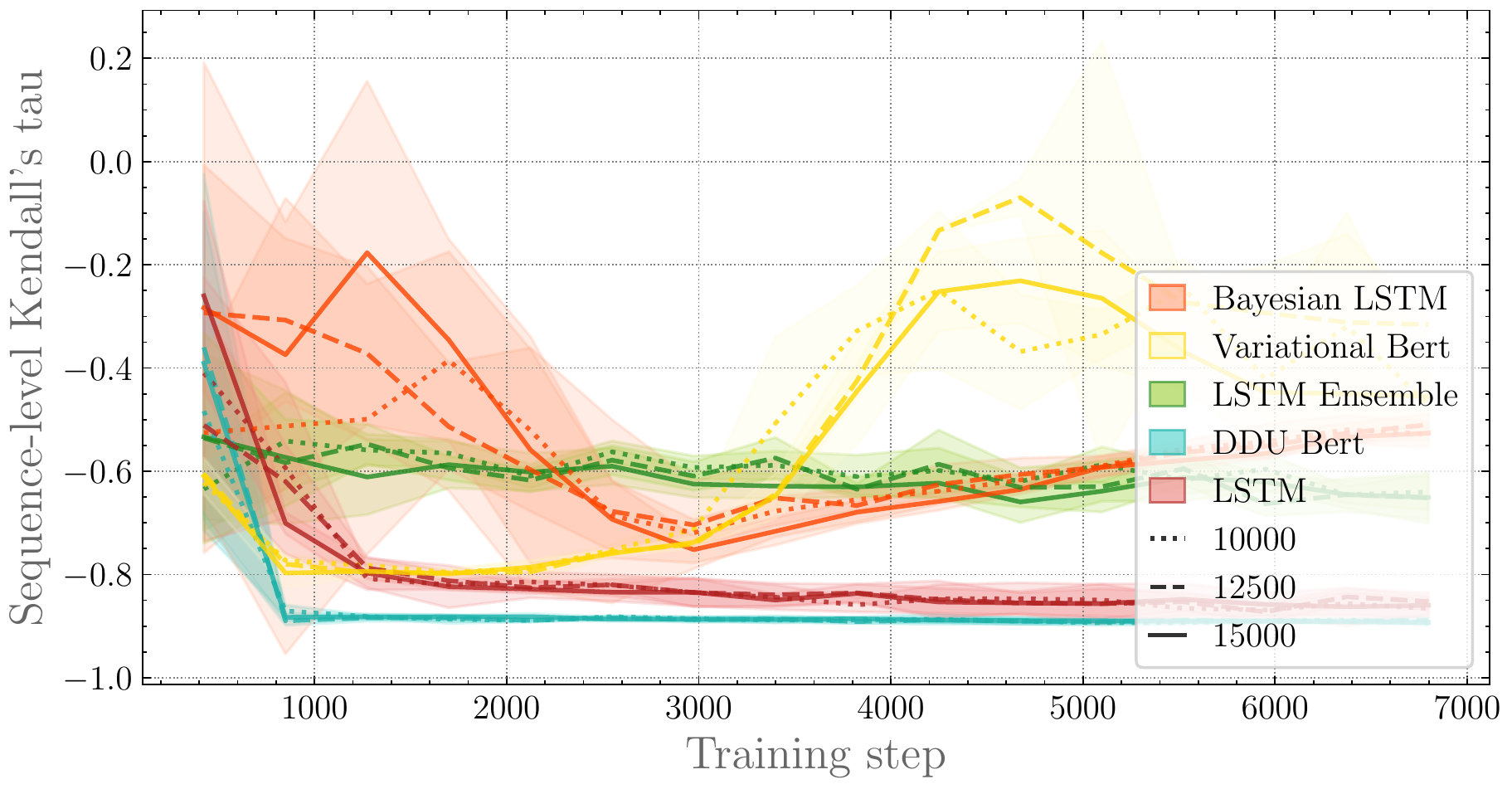}
        \subcaption{Development of sequence-level Kendall's $\tau$ for Clinc Plus.}\label{subfig:development-models-finnish-ud-predictive-entropy-seq}
    \end{subfigure}
    \par\bigskip 
    \begin{subfigure}{\columnwidth}
        \centering
        \includegraphics[width=\columnwidth]{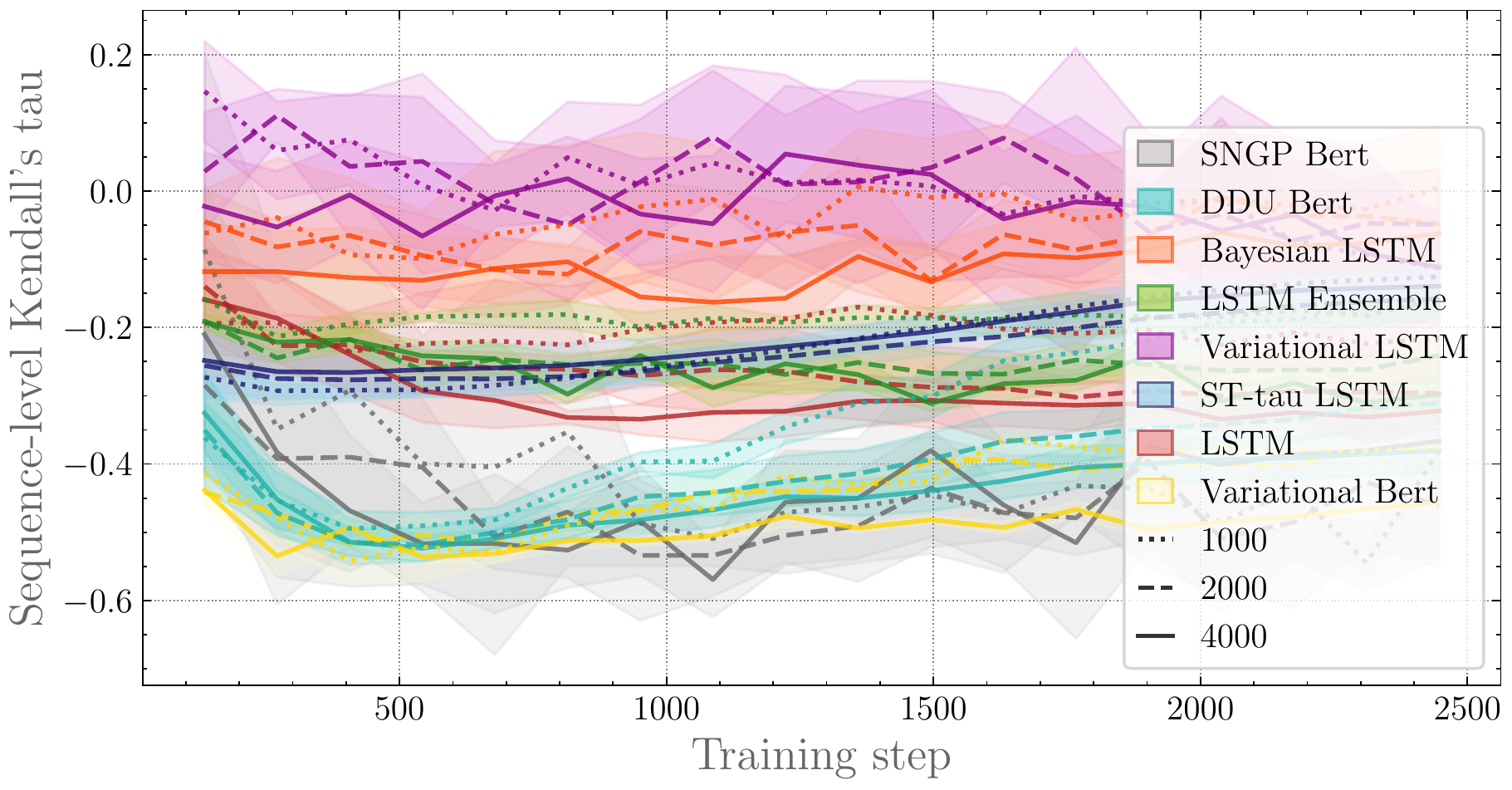}
        \subcaption{Development of sequence-level Kendall's $\tau$ for Dan+.}
    \end{subfigure}\label{subfig:development-models-finnish-ud-predictive-entropy-seq}
    \par\bigskip 
    \begin{subfigure}{\columnwidth}
        \centering
        \includegraphics[width=\columnwidth]{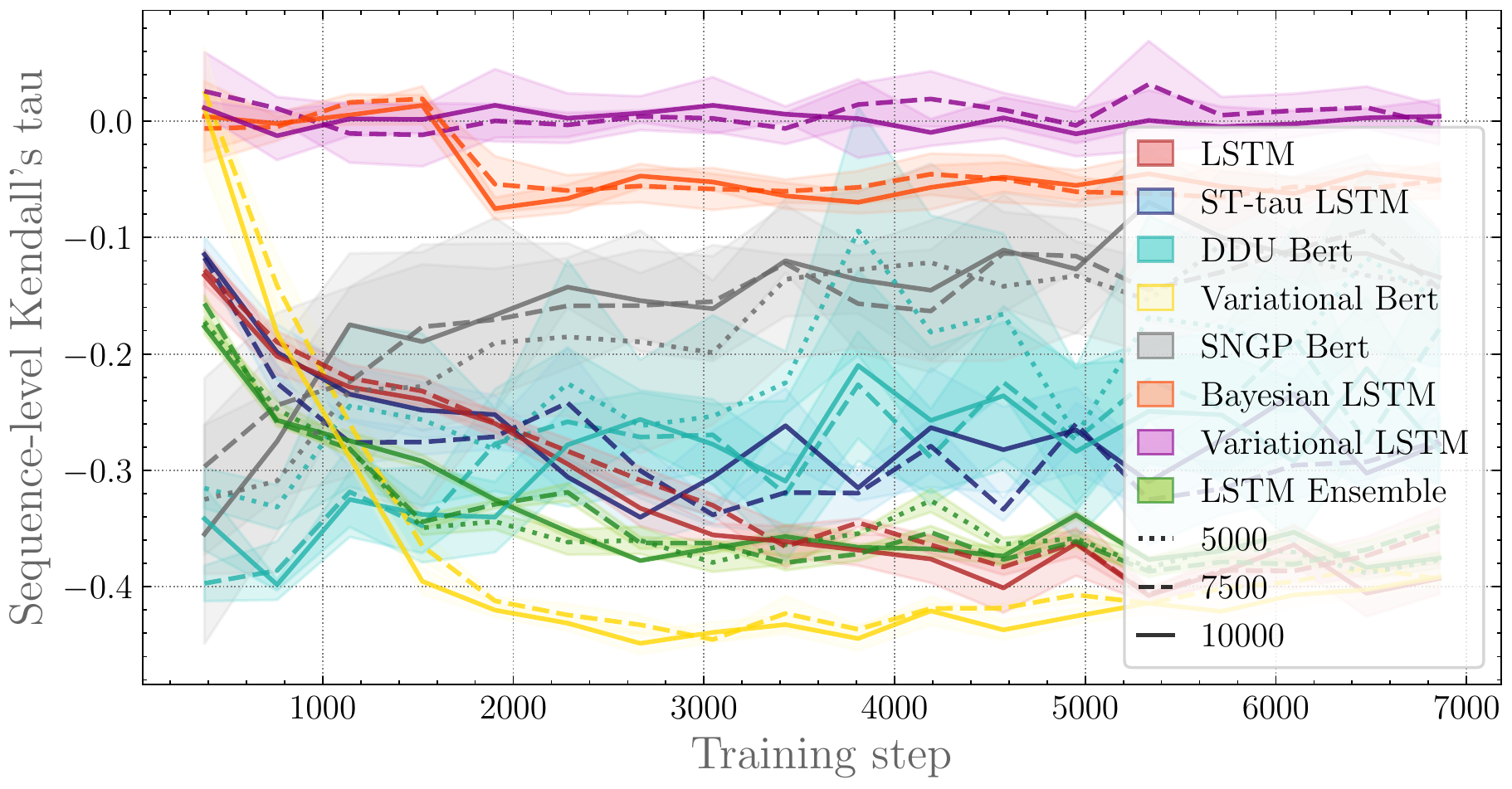}
        \subcaption{Development of sequence-level Kendall's $\tau$ for Finnish UD.}\label{subfig:development-models-finnish-ud-predictive-entropy-seq}
    \end{subfigure}
    \caption{\textbf{Development of correlation between sequence-level predictive entropy and loss on all OOD test sets over the training time}. Data is shown for several model types and using differently-sized training sets. Colored areas indicate the standard deviation over five runs.}\label{fig:development-models-predictive-entropy-sequence-level}
\end{figure}

We extend the plots from \cref{fig:development-lstm-ddu-predictive-entropy} for all tested models and datasets in \cref{fig:development-models-predictive-entropy-token-level} and \cref{fig:development-models-predictive-entropy-sequence-level}, showing the correlation of predictive entropy on a token- and sequence-level, respectively. On the token-level, we see that token-level correlation is the highest for SNGP-BERT, although the correlation levels for training set sizes seems to be harder to differentiate between models and could also be due to variance between models runs. Secondly, on a sequence-level, we also see either similar correlation across training set sizes, or higher correlation for lower sizes. In all cases, we observe that some models start with a high correlation that decreases over the training time, as the model fits the in-distribution data better. That corroborates a trend described in \cref{sec:dependence-training-data}, implying that uncertainty estimates become less reliable as the model tries to decrease the loss on the training data.

\subsection{Qualitative Analysis}\label{app:qualitative analysis}

\begin{figure}[!htb]
    \centering
    \begin{subfigure}[t]{0.99\columnwidth}
        \centering
        \includegraphics[width=\columnwidth]{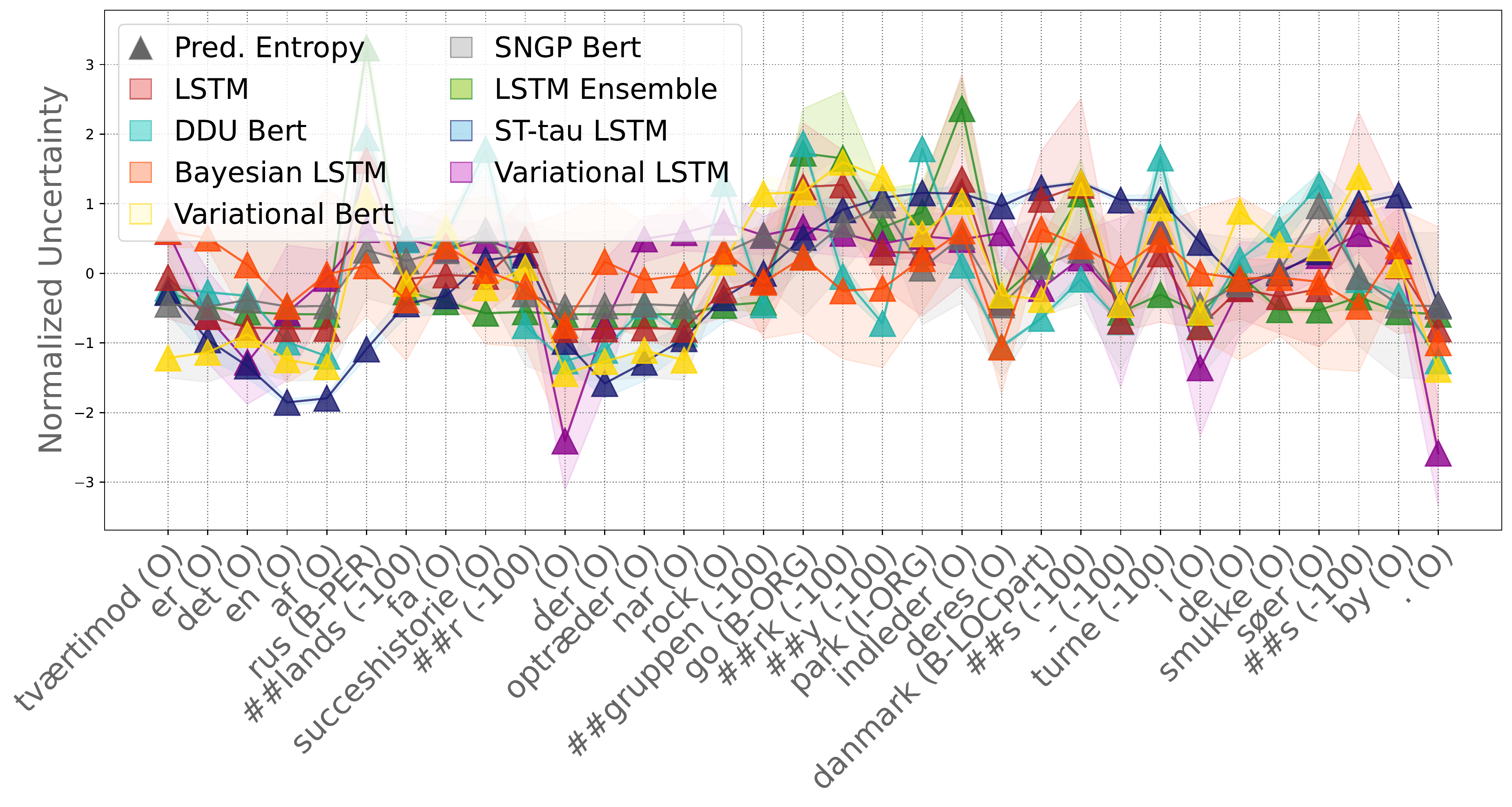}
        \subcaption{Predictive entropy over the sentence \emph{"On the contrary, it is one of Russia's few success stories that performs when the rock group Gorky Park begins their Danish tour in the city of the beautiful lakes"}.}
    \end{subfigure}
    \par\bigskip 
    \centering
    \begin{subfigure}[t]{0.99\columnwidth}
        \centering
        \includegraphics[width=\columnwidth]{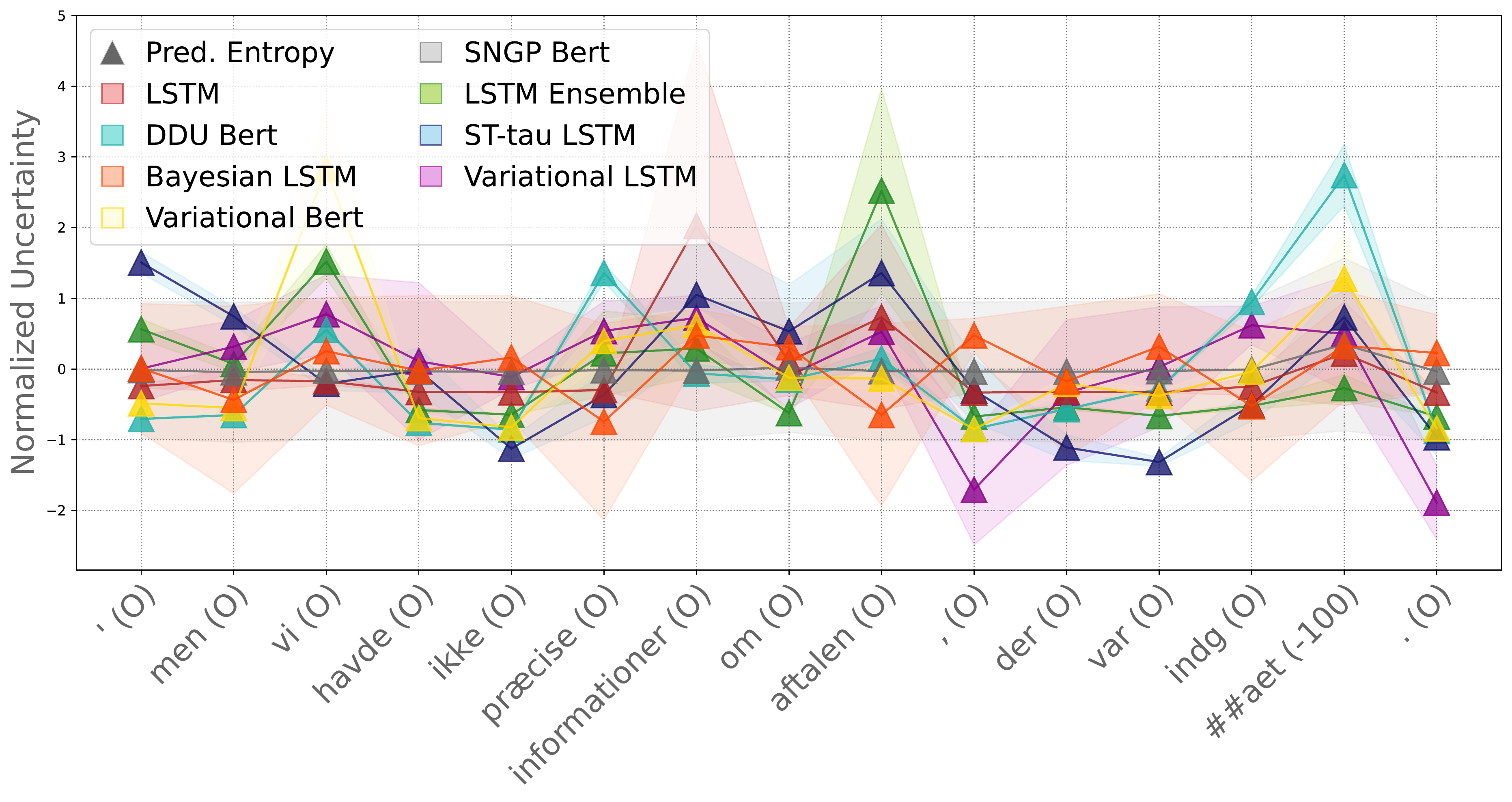}
        \subcaption{Predictive entropy over the sentence \emph{"However, we did not have precise information about what was agreed upon"}.}
    \end{subfigure}
    \par\bigskip 
    \begin{subfigure}[t]{0.99\columnwidth}
        \centering
        \includegraphics[width=\columnwidth]{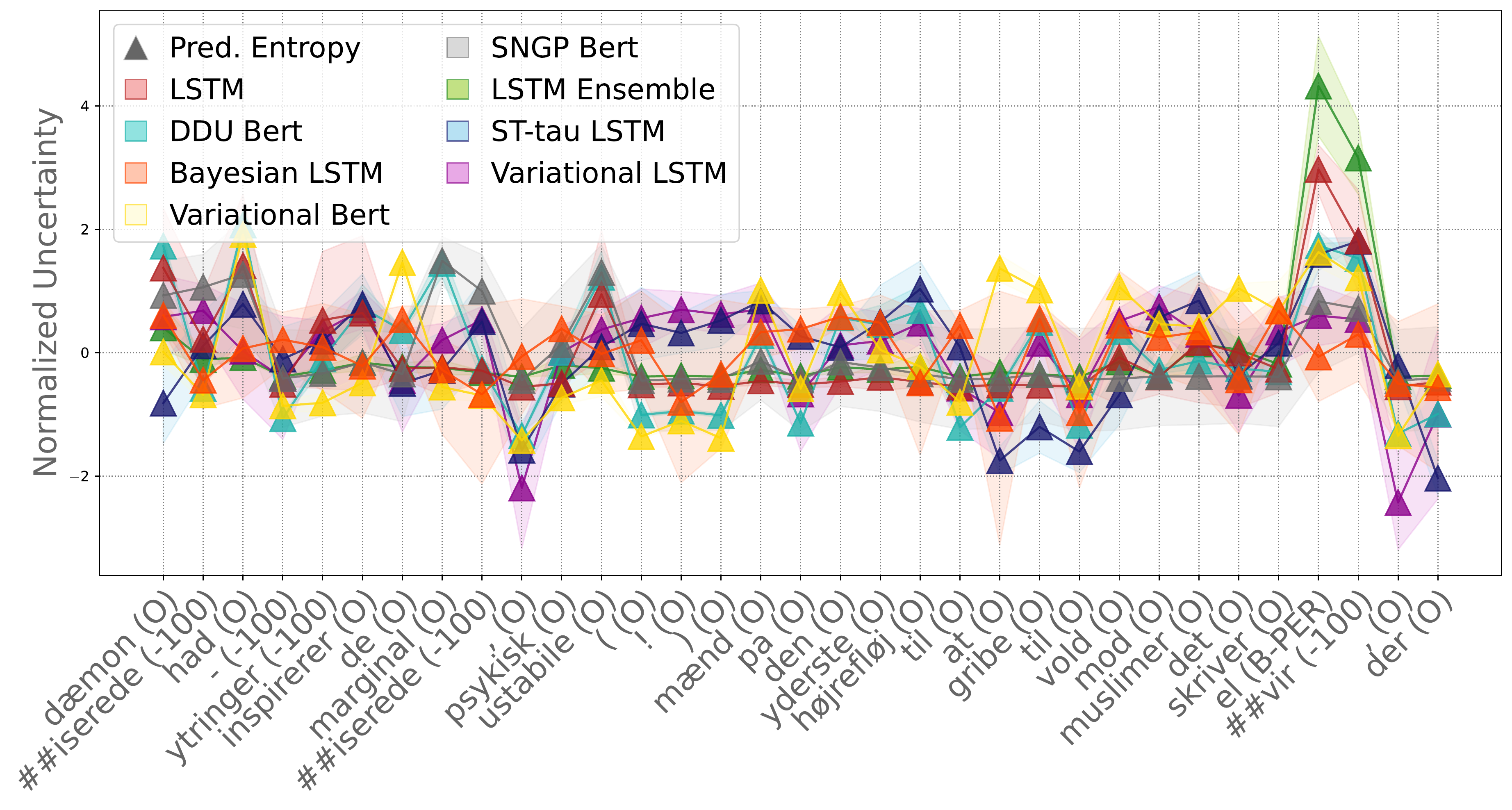}
        \subcaption{Predictive entropy over the sentence \emph{"Demonizing hate speech inspires the marginalized, PSYCHOLOGY UNSTABLE (!) Men on the far right to resort to violence against Muslims. This writes Elvir, who...}.}
    \end{subfigure}
    \caption{
    \textbf{Further examples for uncertainty estimates on single sequences}. Taken from the Dan+ dataset.}\label{fig:qualitative-analysis-extra-danish}
\end{figure}

\paragraph{Dan+} We show more examples of the predictive entropies on samples from the Dan+ dataset in \cref{fig:qualitative-analysis-extra-danish}, where uncertainty values where jointly normalized by subtracting the mean and dividing by the standard deviation over all models and time steps. We can make the following observations: Firstly, uncertainty seems to decrease on punctuation marks such as commas and full-stops. Secondly, uncertainty appears higher on sub-word tokens and some named entities. Thirdly, DDU BERT and the LSTM ensemble produce the highest uncertainty values, which are also two of the best performing models on the task. 

\begin{figure}[!htb]
    \centering
    \begin{subfigure}[t]{0.99\columnwidth}
        \centering
        \includegraphics[width=\columnwidth]{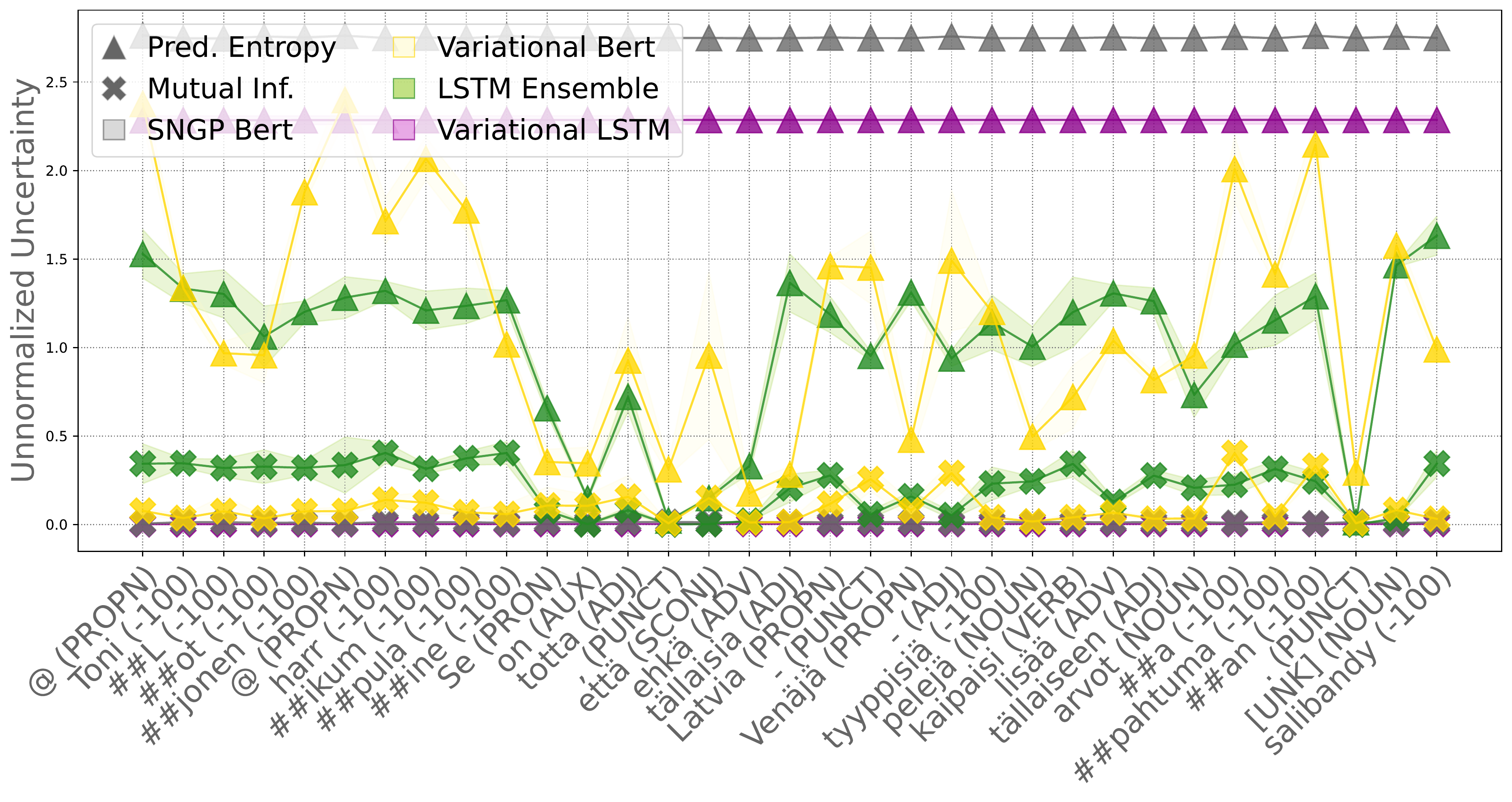}
        \subcaption{Predictive entropy over the sentence \emph{"@ToniLotjonen @harrikumpulaine It is true that I’d maybe like to see more of such Latvia–Russia type games in these kinds of major sports events. \#floorball"}.}
    \end{subfigure}
    \par\bigskip 
    \centering
    \begin{subfigure}[t]{0.99\columnwidth}
        \centering
        \includegraphics[width=\columnwidth]{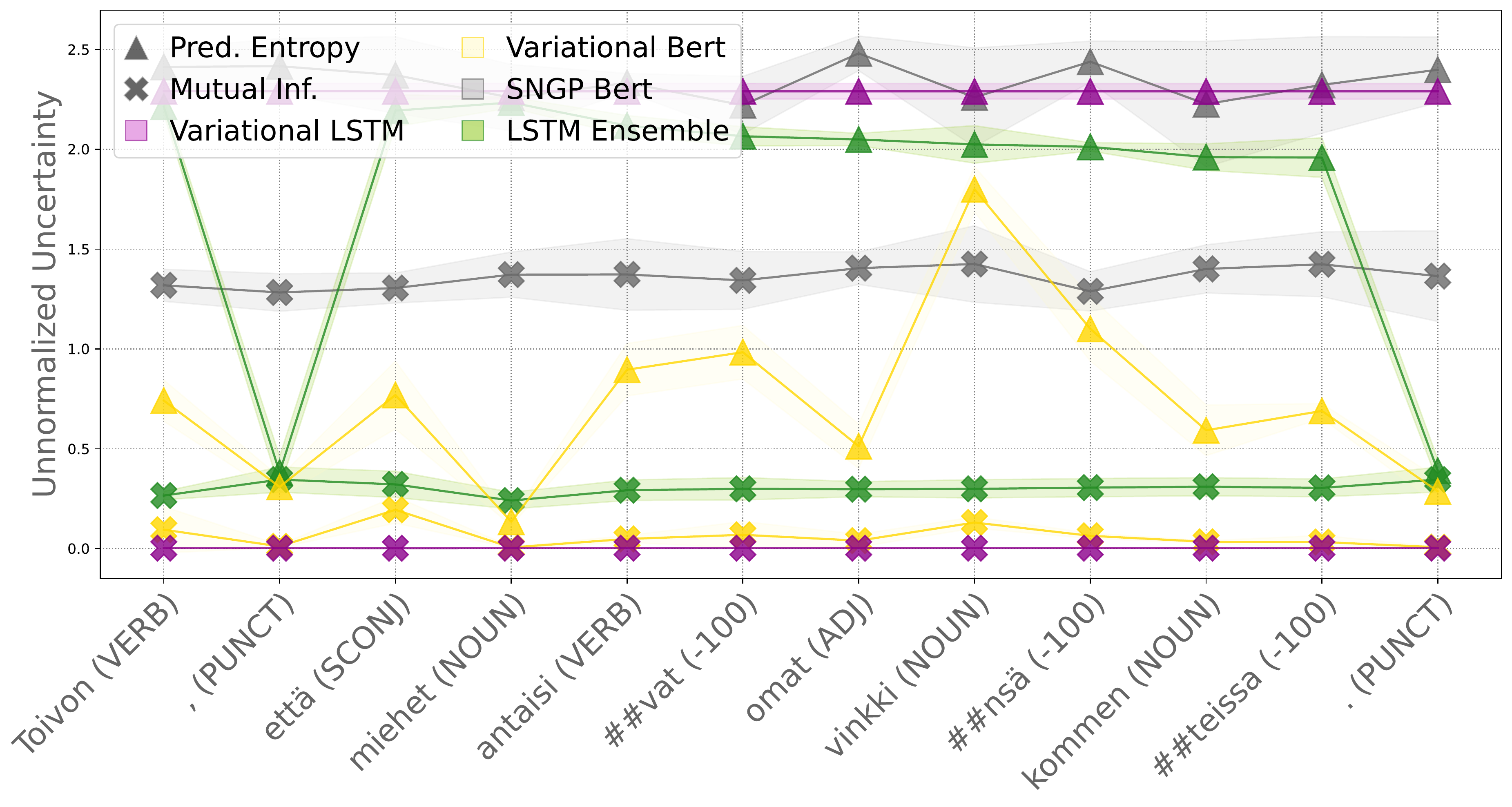}
        \subcaption{Predictive entropy over the sentence \emph{"I hope that the procedures done on the  person in question stop and he gives his body (and mind) time to recover from that poisoning!"}.}
    \end{subfigure}
    \par\bigskip 
    \begin{subfigure}[t]{0.99\columnwidth}
        \centering
        \includegraphics[width=\columnwidth]{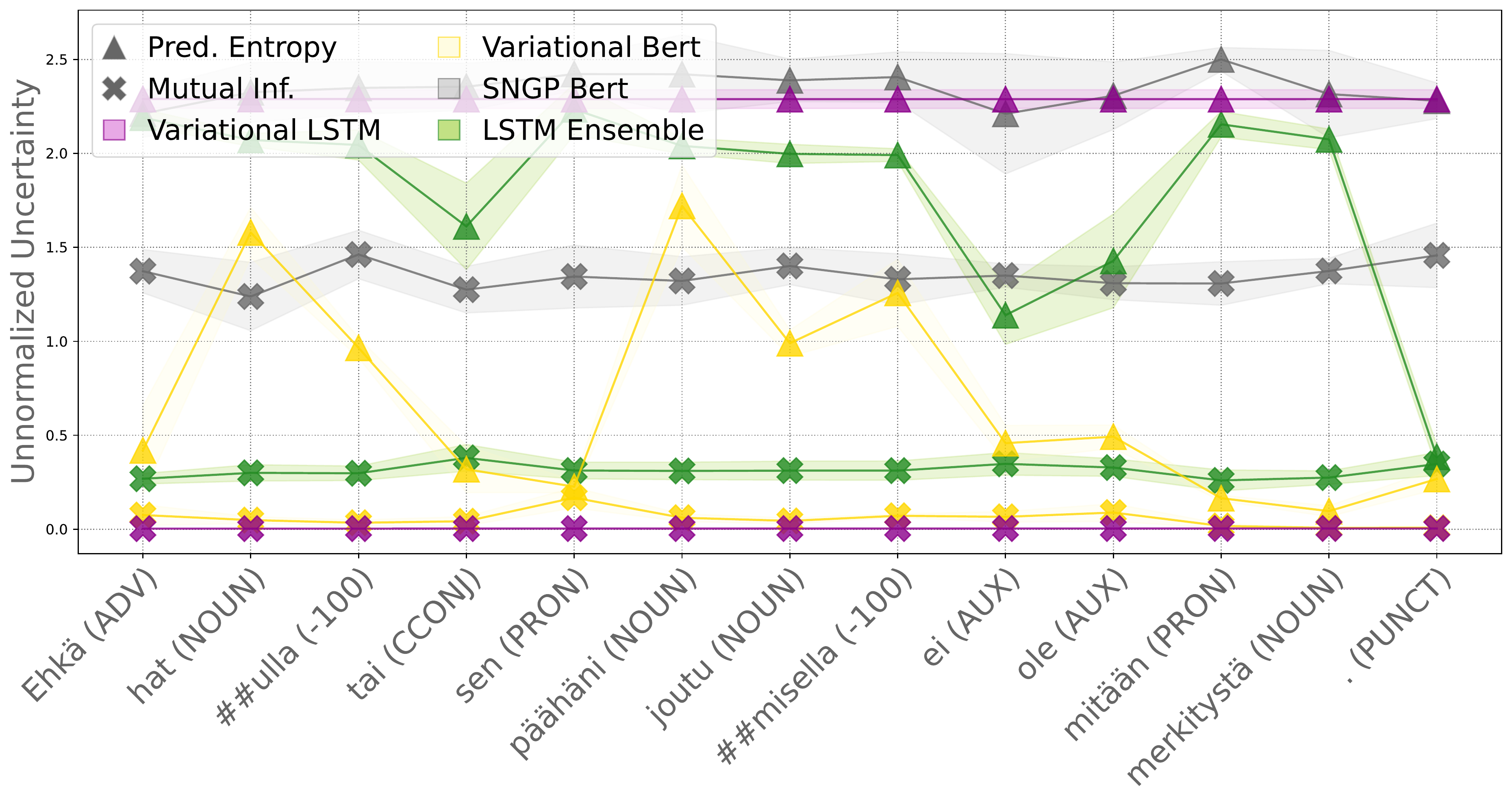}
        \subcaption{Predictive entropy over the sentence \emph{"Maybe the hat or how it got on my head doesn’t matter"}.}
    \end{subfigure}
    \caption{\textbf{Further examples for uncertainty estimates on single sequences}. Taken from the Finnish UD dataset.}\label{fig:qualitative-analysis-extra-finnish}
\end{figure}

\paragraph{Finnish UD} Here, we give more examples of the analysis on the Finnish UD dataset in \cref{fig:qualitative-analysis-extra-finnish}. First of all, we see that the Variational LSTM and SNGP BERT seem to produce almost constant uncertainty scores, which can be explained by their suboptimal performance in task, as shown by their results in \cref{table:results}. But even for the models that perform better, such as the Variational BERT and the LSTM ensemble, the decomposition of predictive entropy into aleatoric and epistemic uncertainty reveals that model uncertainty generally remains low, and is overshadowed to a larger extent by the aleatoric uncertainty. We can observe that similar to Danish, uncertainty seems to be low on punctuation marks and high on subword tokens. Furthermore, aleatoric uncertainty seems to be higher on nouns and pronouns. This could be due to the sheer number of possible nouns and pronouns that could fill such a gap in a sentence.

\end{document}